\documentclass[11pt]{article}

\usepackage[preprint]{acl}

\usepackage{times}
\usepackage{latexsym}

\usepackage[T1]{fontenc}

\usepackage[utf8]{inputenc}

\usepackage{microtype}

\usepackage{inconsolata}

\usepackage{graphicx}
\usepackage{amsmath}

\title{
    \makebox[0pt][r]{%
        \raisebox{-0.25\height}{\includegraphics[width=2em]{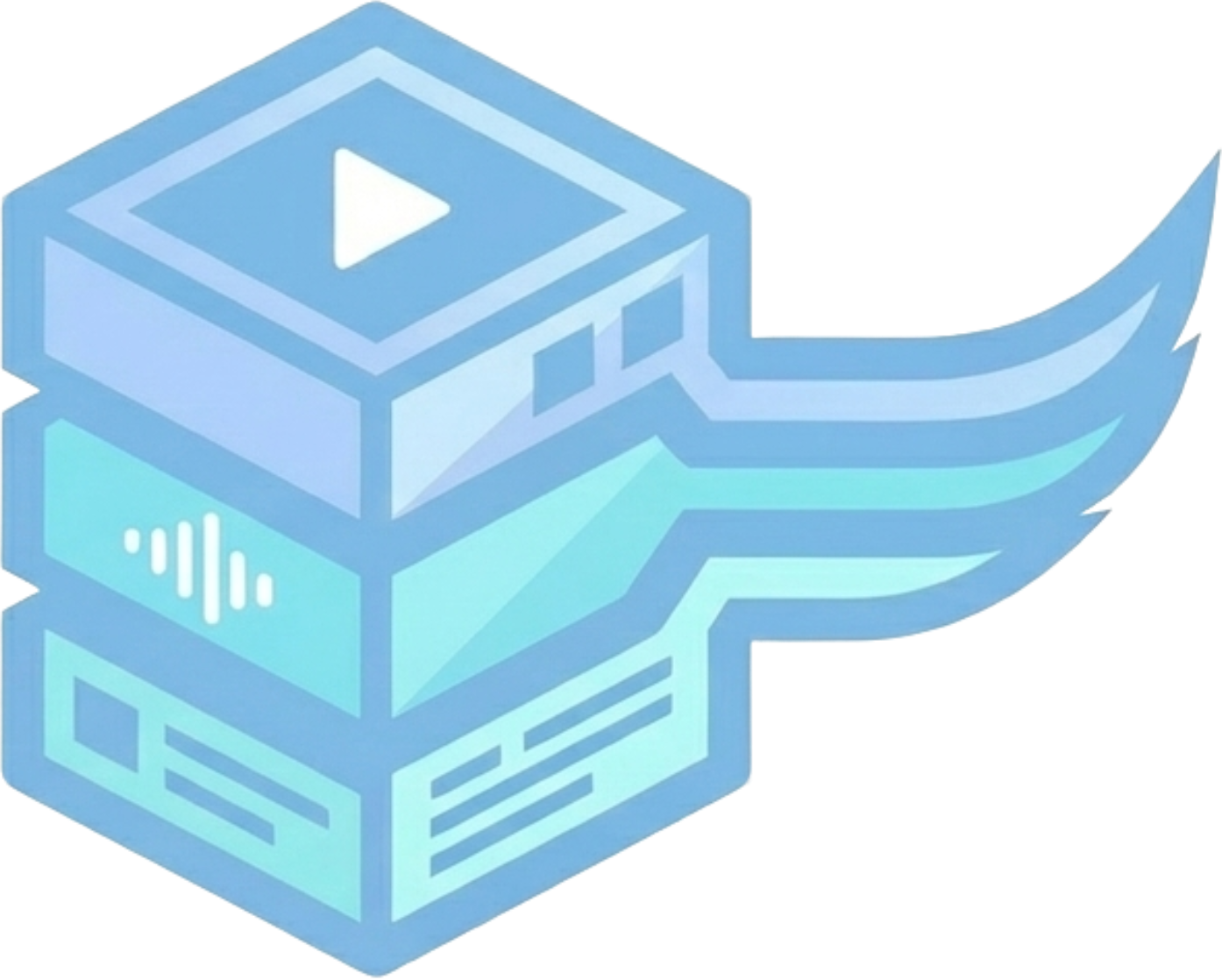}}%
        \hspace{0.25em}
    }%
    ROMA: Real-time Omni-Multimodal Assistant \\
    with Interactive Streaming Understanding
}

\author{
    Xueyun Tian$^{\spadesuit\heartsuit}$, 
    Wei Li, 
    Bingbing Xu$^{\spadesuit}$, 
    Heng Dong$^{\clubsuit}$, 
    Yuanzhuo Wang$^{\spadesuit}$, 
    Huawei Shen$^{\spadesuit\heartsuit}$ \\
    $^{\spadesuit}$CAS Key Laboratory of AI Safety, Institute of Computing Technology, CAS, Beijing, China \\
    $^{\heartsuit}$University of Chinese Academy of Sciences, Beijing, China \\
    $^{\clubsuit}$Tsinghua University, Beijing, China \\
    \small \texttt{\{tianxueyun23z, xubingbing, wangyuanzhuo, shenhuawei\}@ict.ac.cn} \\
    \small \texttt{weili.ucas.ict@gmail.com}, \texttt{drdhxi@gmail.com}
}

\usepackage{array}
\usepackage{tabularray}
\usepackage{amsmath,amssymb}
\usepackage{hyperref}
\usepackage{booktabs}  
\usepackage{tabularx}  
\renewcommand{\tabularxcolumn}[1]{m{#1}} 
\usepackage{xcolor}    
\usepackage{colortbl}  
\usepackage{pifont}    
\usepackage{threeparttablex}
\definecolor{softblue}{RGB}{235, 242, 250}
\definecolor{grpA}{HTML}{F7FAFC} 
\definecolor{grpB}{HTML}{FAF6FB} 
\definecolor{grpC}{HTML}{F6FBF7} 
\definecolor{grpD}{HTML}{FCF7F1} 
\definecolor{grpE}{HTML}{F5FAFA} 
\definecolor{grpOurs}{HTML}{FFF6D8} 
\newcolumntype{Y}{>{\raggedright\arraybackslash}X}
\newcolumntype{C}[1]{>{\centering\arraybackslash}m{#1}}
\usepackage{ragged2e}
\newcolumntype{L}[1]{>{\RaggedRight\arraybackslash}m{#1}}

\definecolor{softgreen}{RGB}{85, 160, 85}
\definecolor{softred}{RGB}{180, 90, 90}
\newcommand{\yes}{\textcolor{softgreen}{\ding{51}}} 
\newcommand{\no}{\textcolor{softred}{\ding{55}}}   

\usepackage{xcolor}
\usepackage{listings}
\usepackage[most]{tcolorbox}
\tcbuselibrary{listings,skins}

\definecolor{PromptBg}{HTML}{F6FAFC}        
\definecolor{PromptFrame}{HTML}{A7B4C2}     
\definecolor{PromptTitleBg}{HTML}{6F7F8F}   
\definecolor{PromptTitleFg}{HTML}{FFFFFF}

\lstdefinestyle{PromptStyle}{
  basicstyle=\ttfamily\footnotesize, 
  breaklines=true,                 
  breakatwhitespace=false,         
  columns=fullflexible,            
  keepspaces=true,
  showstringspaces=false,
  frame=none,                      
  aboveskip=0pt,
  belowskip=0pt
}

\newtcolorbox{promptbox}[2][]{%
  enhanced,
  colback=PromptBg,
  colframe=PromptFrame,
  boxrule=0.6pt,
  arc=2mm,
  left=2.2mm,right=2.2mm,top=1.2mm,bottom=1.2mm,
  fonttitle=\bfseries,
  colbacktitle=PromptTitleBg,
  coltitle=PromptTitleFg,
  boxed title style={boxrule=0pt,arc=2mm},
  attach boxed title to top left={xshift=2mm,yshift=-1mm},
  title={#2},
  #1
}

\definecolor{tblHeader}{HTML}{1F4E79} 
\definecolor{tblOurs}{HTML}{DCEBFA}   
\definecolor{tblSection}{HTML}{F3F7FC}
\definecolor{tblRule}{HTML}{4B5563}   
\definecolor{tblRuleLight}{HTML}{9CA3AF} 
\usepackage{xspace}
\newcommand{\ours}{\textsc{ROMA}\xspace}
\begin{document}
\maketitle
\begin{abstract}
Recent Omni-multimodal Large Language Models show promise in unified audio, vision, and text modeling. However, streaming audio-video understanding remains challenging, as existing approaches suffer from disjointed capabilities: they typically exhibit incomplete modality support or lack autonomous proactive monitoring. 
To address this, we present \ours, \textbf{a real-time omni-multimodal assistant for unified reactive and proactive interaction.} \ours processes continuous inputs as synchronized \textit{multimodal units}, aligning dense audio with discrete video frames to handle granularity mismatches. 
For online decision-making, we introduce a lightweight \textit{speak head} that decouples response initiation from generation to ensure precise triggering without task conflict. 
We train \ours with a curated streaming dataset and a two-stage curriculum that progressively optimizes for streaming format adaptation and proactive responsiveness.
To standardize the fragmented evaluation landscape, we reorganize diverse benchmarks into a unified suite covering both proactive (alert, narration) and reactive (QA) settings. Extensive experiments across 12 benchmarks demonstrate \ours achieves state-of-the-art performance on proactive tasks while competitive in reactive settings, validating its robustness in unified real-time omni-multimodal understanding.
Our project page is available at \href{https://eureka-maggie.github.io/ROMA_show/}{here}\footnote{\url{https://eureka-maggie.github.io/ROMA_show/}}.


\end{abstract}
\begin{figure}[ht]
  \centering
  \includegraphics[width=\linewidth]{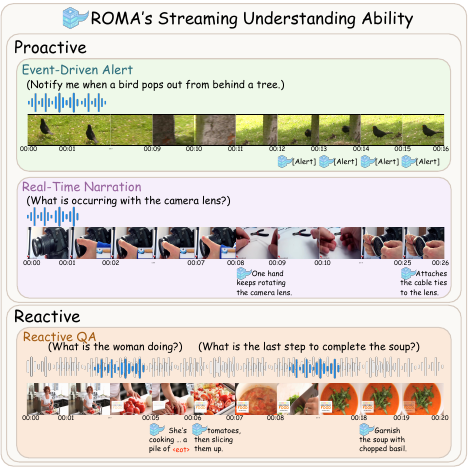}
  \caption{\textbf{ROMA's streaming understanding capabilities.} It supports proactive tasks, including event alerts and narration, alongside reactive question answering.}
  \label{fig:task}
\vspace{-11pt}
\end{figure}
\section{Introduction}
Recent advances in omni-multimodal large language models (OLLMs), such as GPT-4o~\cite{hurst2024gpt}, have enabled unified modeling of speech, vision, and text. This progress facilitates real-world streaming audio-video understanding, defined as combining \textbf{reactive} and \textbf{proactive} capabilities (Figure~\ref{fig:task}). In the reactive setting, the model answers after the query, whereas in the proactive setting, it follows an instruction to continuously monitor the input stream and respond only when conditions are met. 
Unifying these capabilities is vital for real-world utility, yet the divergent interaction paradigms make it challenging~\cite{horvitz1999principles, xi2025rise,driess2023palm}.

Despite the critical need for such unification, existing studies typically lack unified modality support and streaming capabilities. 
Specifically, speech-centric streaming models~\cite{defossez2024moshi,zhang2025stream} focus on audio generation but lack visual perception. Conversely, while some approaches address streaming video understanding~\cite{chen2024videollm,zhang2024internlm}, they typically neglect synchronized audio and are confined to specific tasks (e.g., alert or narration).
Consequently, unified streaming audio-video understanding remains largely under-explored.


To realize such unification faces two challenges.
First, audio and video exhibit mismatched temporal granularities. While naturally synchronized, audio signals are dense and continuous, whereas video comprises sparse, discrete frames. Under such heterogeneity, maintaining robust cross-modal alignment and fusion demands precise synchronization. 
Second, effective streaming interaction requires real-time proactive decision-making. Upon integrating these asynchronous signals, the model must continuously synthesize context to determine both response timing and content, conditioned strictly on the stream prefix.
To address these challenges, we propose \ours, a \textbf{R}eal-time \textbf{O}mni-\textbf{M}ultimodal \textbf{A}ssistant with interactive streaming understanding. 
To tackle the granularity mismatch, \ours segments continuous audio into one-second intervals synchronized with video frames, forming temporally aligned units that are processed sequentially as the stream unfolds.
We further adapt chunked Time-aligned Multimodal RoPE (TMRoPE)~\cite{xu2025qwen2} to enforce a shared temporal timeline. 
For proactive decision-making, \ours introduces a lightweight speak head parallel to the standard language modeling (LM) head to explicitly predict response timing, decoupling timing from content generation to prevent task interference.
Finally, we support this system with a custom streaming dataset and a two-stage training curriculum, progressively optimizing the model for cross-modal streaming format adaptation and proactive responsiveness.


For a comprehensive evaluation, streaming audio-video understanding demands assessing both reactive and proactive capabilities. However, as compared in Table~\ref{tab:related_bench}, existing benchmarks suffer from inconsistent taxonomies and fragmented protocols, often failing to cover both interaction modes. To enable unified comparison, we reorganize the evaluation landscape into two standardized settings: a \textit{proactive} mode that tests the ability to autonomously trigger responses at precise moments, and a \textit{reactive} mode that emphasizes understanding temporal evolution in standard QA. Empirically, \ours consistently outperforms existing streaming VideoLLMs across both modes. Furthermore, evaluations on open-ended audio-query QA against open-source OLLMs confirm its superior capability in unified audio-video understanding.

In summary, our contributions are as follows:
\begin{itemize}
  \item \textbf{Unified streaming framework:} We formally define the task of streaming audio-video understanding and propose \ours, an omni-multimodal assistant unifying reactive and proactive capabilities, supported by a curated dataset and a two-stage curriculum.
  \item \textbf{Standardized evaluation benchmark:} We establish a comprehensive streaming benchmark by reorganizing fragmented tasks into unified \textit{reactive} and \textit{proactive} settings to facilitate rigorous and consistent comparison.
  \item \textbf{Superior performance and analysis:} \ours achieves state-of-the-art results across proactive benchmarks while competitive on reactive and open-ended QA. Extensive analysis verifies the efficacy of our timing mechanisms and training strategies.

\end{itemize}

\newcommand{\benchcite}[2]{%
  \begin{tabular}[t]{@{}l@{}}#1\\[-2pt]{\footnotesize\cite{#2}}\end{tabular}%
}
\begin{table}[htbp]
\vspace{-5pt}
\centering
\small
\setlength{\tabcolsep}{4pt}
\renewcommand{\arraystretch}{1}

\begin{tabularx}{\linewidth}{X c c c}
\toprule
\rowcolor{softblue}
\textbf{Benchmark} & \textbf{Alert} & \textbf{Narration} & \textbf{Reactive QA} \\
\midrule
\benchcite{StreamingBench}{lin2024streamingbench} & \yes & \no  & \yes \\
\benchcite{StreamBench}{wu2024streambench}  & \no  & \no  & \yes \\
\benchcite{OVO-Bench}{niu2025ovo}   & \no  & \yes & \yes \\
\benchcite{SVBench}{yang2025svbench}    & \no  & \no  & \yes \\
\benchcite{OmniMMI}{wang2025omnimmi}  & \yes & \no  & \yes \\
\benchcite{OVBench}{huang2025online}  & \no  & \no  & \yes \\
Ours & \yes & \yes & \yes \\
\bottomrule
\end{tabularx}

\caption{Coverage of key streaming ability across representative streaming video benchmarks.}
\label{tab:related_bench}
\vspace{-10pt}
\end{table}

\begin{figure*}[t]
  \centering
  \includegraphics[width=\linewidth]{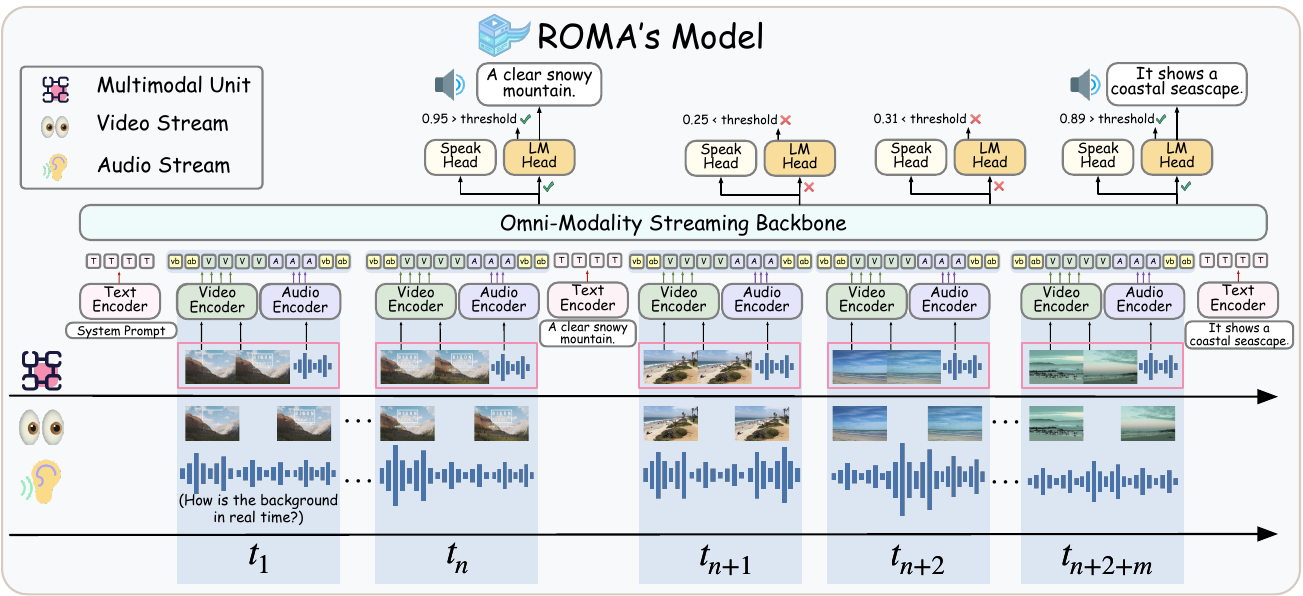}
  \caption{\textbf{Model Architecture.} Streaming inputs are processed as aligned multimodal units. The speak head determines response timing, activating the LM head (illustrated via narration) upon crossing a probability threshold.}
  \label{fig:model}
\vspace{-10pt}
\end{figure*}

\section{Related Works}
\paragraph{Reactive Models}
Most existing streaming systems are studied in the reactive setting, answering only after the query arrives. Within this regime, memory-based methods maintain long-range context for coherent understanding over evolving streams~\cite{qian2024streaming,zhang2024flash,wang2024videollamb,zhang2024internlm,xiong2025streaming,wang2025streambridge,zhao2025cogstream}, and KV-cache based methods optimize efficiency via scheduling or compression~\cite{di2025streaming,ning2025livevlm,yang2025streamagent,xu2025streamingvlm,chen2025streamkv}. Recent omni-multimodal models also adhere to this reactive protocol: MiniCPM-o 2.6~\cite{yao2024minicpm}, Qwen2.5-Omni~\cite{xu2025qwen2}, and Qwen3-Omni~\cite{Qwen3-Omni} support low-latency interaction, and Stream-Omni~\cite{zhang2025stream} enables visually-conditioned speech generation, yet none explicitly model proactive monitoring and triggering.


\paragraph{Proactive Models}
In contrast, proactive streaming prioritizes continuous monitoring and time-sensitive triggering (e.g., alerts and real-time narration). Proactive VideoLLMs leverage online formats or explicit decision modeling to determine intervention timing~\cite{chen2024videollm,yang2025assistpda,li2025lion,yang2025livestar,qian2025dispider}, with some explicitly targeting live narration~\cite{chen2025livecc}. However, these approaches remain predominantly video-centric, neglecting streaming audio.


\paragraph{Streaming Video Understanding Benchmarks}
Recent benchmarks prioritize time-sensitive, interactive evaluation (Table~\ref{tab:related_bench}). StreamingBench~\cite{lin2024streamingbench} and OVO-Bench~\cite{niu2025ovo} assess temporal perception, while StreamBench~\cite{wu2024streambench} and SVBench~\cite{yang2025svbench} focus on long-horizon memory. Moreover, OmniMMI~\cite{wang2025omnimmi} and OVBench~\cite{huang2025online} incorporate proactive capabilities, including real-time narration and alerts.

\textbf{Tables~\ref{tab:related_bench} and~\ref{tab:related_work} summarizes prior works.}

\section{Method}
To unify reactive answering and proactive timing over continuous inputs, \ours integrates architectural designs with a tailored training strategy. Section~\ref{sec:arch} presents the model architecture, utilizing chunked TMRoPE and a speak head for alignment and timing control. Section~\ref{sec:pipeline} details the training and inference pipeline, encompassing dataset construction and a two-stage fine-tuning recipe.

\subsection{Model Architecture}
\label{sec:arch}
As illustrated in Figure~\ref{fig:model}, \ours processes streaming omni-modal inputs via a unified LLM backbone. We introduce a speak head parallel to the LM head to decouple interaction timing from content generation. This architecture addresses temporal alignment and proactive decision-making through the following mechanisms.


\paragraph{Multimodal units for temporally aligned streaming inputs.}
To support unified streaming understanding across modalities, we organize audio and video into fixed-interval multimodal units. Following the input format and tokenization of Qwen2.5-Omni, we treat all audio and video signals within each one-second interval as a unit. We align audio with video frames sampled from the same interval, extract their features, and wrap them with special tokens. This retains Qwen2.5-Omni's native format for compatibility while grounding audio in the preceding visual context:
\newcommand{\tok}[1]{\texttt{<|#1|>}}
\setlength{\fboxsep}{2pt}
\setlength{\fboxrule}{0.4pt}

\begin{center}
\fbox{%
\parbox{0.92\columnwidth}{\centering\footnotesize
\tok{vision\_bos} \tok{audio\_bos} [\textit{video tokens}] [\textit{audio tokens}]
\tok{audio\_eos} \tok{vision\_eos}%
}}
\end{center}

These multimodal units are fed into the LLM backbone sequentially as the stream unfolds. This process ensures that the model continuously accumulates aligned cross-modal context from the stream prefix, establishing a temporal basis for subsequent causal decision-making.

\paragraph{Chunk-Level Temporal Position Encoding}
We adapt Qwen2.5-Omni's Time-aligned Multimodal RoPE (TMRoPE) to chunked audio–video streams to support incremental encoding as units arrive. Each one-second unit interleaves visual and auditory tokens, assigning time-aligned 3D position IDs (temporal, height, and width) to preserve their cross-modal correspondence. 
Consistent with the pre-trained vision encoder, multi-frame visual inputs are temporally aggregated into a fused representation during encoding. All video tokens within a unit therefore share a constant temporal ID. In contrast, audio tokens retain fine-grained temporal IDs at a $40\text{ms}$ resolution to preserve auditory temporal fidelity. To ensure boundary alignment, \tok{vision\_bos} and \tok{audio\_bos} share the same base position ID, subsequent units extend the global timeline by continuing from the maximum position ID of the previous unit.

\begin{figure}[h]
\vspace{-2pt}
  \centering
  \includegraphics[width=\linewidth]{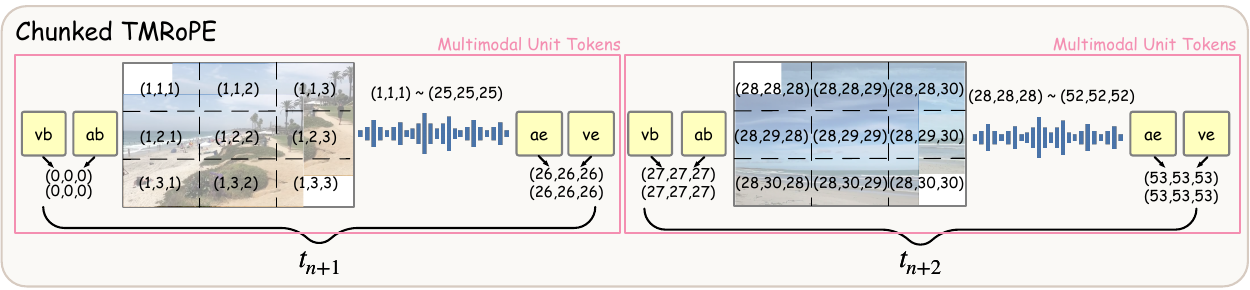}
  \caption{\textbf{Chunked TMRoPE.} Seamlessly extends the global timeline to streaming inputs by assigning cumulative positional IDs across discrete units.}
  \label{fig:tmrope}
  \vspace{-8pt}
\end{figure}

\paragraph{Speak Head}
To enable autonomous intervention timing, we design a lightweight speak head. As illustrated in Figure~\ref{fig:model}, this module is implemented as a two-layer MLP, parallel to the LM head, on top of the streaming backbone. Upon processing each multimodal unit (one second of context), the speak head evaluates the current stream prefix and outputs a probability for a binary decision indicating whether a response is required. A response is triggered if this probability exceeds a threshold; otherwise, the model remains silent and continues consuming the stream. This design decouples the timing decision from text generation, mitigating interference from generative biases. Leveraging findings that upper layers encode high-level features~\cite{tenney2019bert,belrose2023eliciting}, we compute the speak head input as a learnable weighted combination of hidden states from the last $K$ layers, with $K{=}4$ in our experiments.

\subsection{Training and Inference Pipeline}
\label{sec:pipeline}
\subsubsection{Dataset Construction}
To enable end-to-end proactive and reactive supervision, we construct a comprehensive streaming dataset structured into two categories and three sub-tasks (Figure~\ref{fig:dataset}). Detailed processing pipelines are provided in Appendix~\ref{app:data}.

\paragraph{Online Proactive (27K)} To equip the model with the ability to continuously monitor streams and trigger alerts, we curate data from DiDeMo~\cite{anne2017localizing}, OOPS~\cite{epstein2020oops}, and Charades-STA~\cite{zhou2018towards}. We reformulate these samples into alert-style tasks (e.g., ``Alert me when [event] happens'') to train the model in event-driven temporal grounding.

\paragraph{Online Narration (109K)} To foster continuous event tracking and incremental summarization, we construct narration samples from MMDuetIT~\cite{wang2024videollm}, COIN~\cite{tang2019coin}, YouCook2~\cite{zhou2018towards}, and ActivityNet~\cite{caba2015activitynet}. Unlike prior works that use dense supervision, we specifically train the model to generate captions only at segment transitions, enabling it to provide concise, real-time updates as the visual context evolves.

\paragraph{Reactive QA (540K)} To stabilize general audio–video understanding, we aggregate large-scale reactive QA data from InternVid~\cite{wang2023internvid}, CogStream~\cite{zhao2025cogstream}, and others~\cite{chen2023egoplan,yang2022avqa,yao2025timechat,fu2025vispeak}. These samples cover past events, temporal ordering, and future reasoning.

To ensure unified processing, we synthesize text queries into speech, training the model to handle audio instructions under streaming inputs.

\begin{figure}[htbp]
\vspace{-2pt}
  \centering
  \includegraphics[width=\columnwidth]{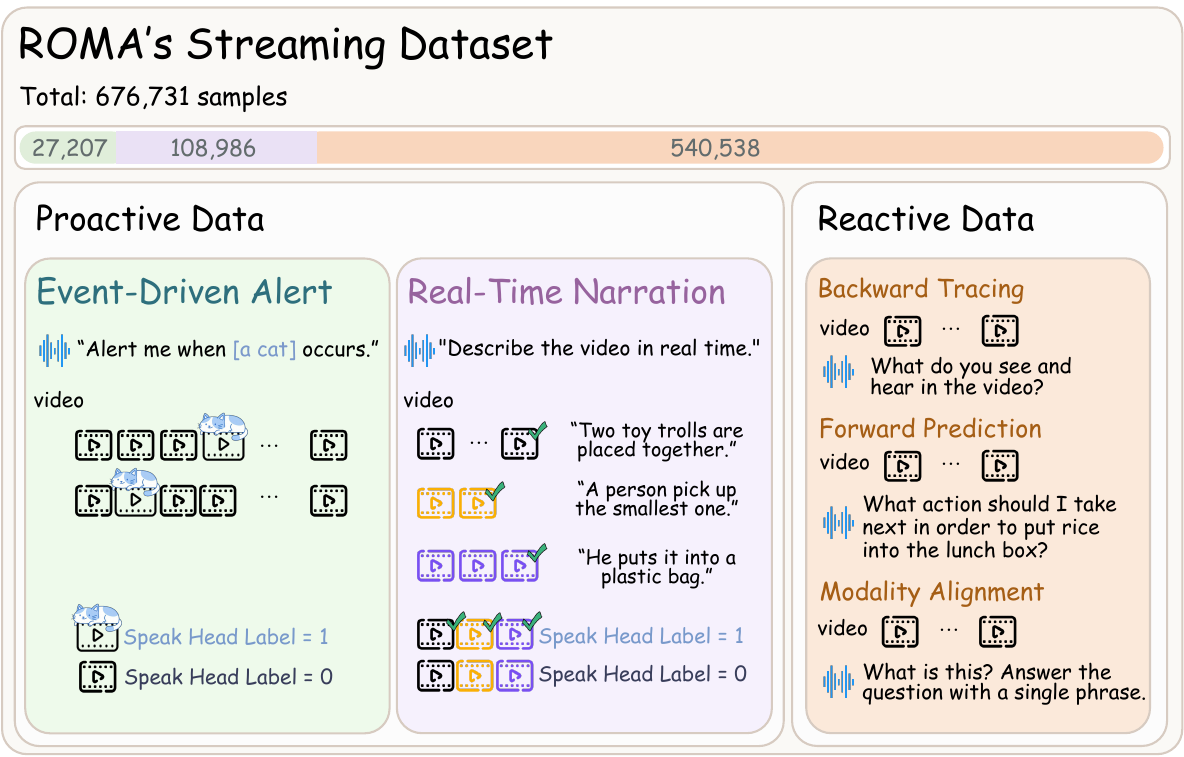}
  \caption{\textbf{Overview of ROMA's Streaming Dataset.} Statistics, task taxonomy, and sample formats.}
  \label{fig:dataset}
  \vspace{-8pt}
\end{figure}
\subsubsection{A Two-Stage Fine-Tuning Recipe}
Training an end-to-end streaming omni-multimodal model from scratch is computationally prohibitive. We fundamentally view streaming capability as a transfer problem: adapting a strong foundation model optimized for processing complete videos to handle incremental streams. We thus propose a simple yet effective two-stage recipe. Stage 1 adapts the model to the streaming multimodal input format, while Stage 2 learns precise response timing and proactive policies. In both stages, we freeze all encoders and fine-tune the remaining parameters $\theta$.

\paragraph{Stage 1: Streaming Template Alignment}
This stage mitigates the distribution shift between offline training and streaming inference. We utilize reactive QA datasets to adapt the model to the multimodal unit streaming format. Samples are restructured into sequential units $X$ to simulate streaming, with the audio query and text response $Y$ appended.
We optimize the standard autoregressive language modeling objective over the response tokens. Let $\mathcal{D}_{\text{QA}}$ denote the reactive QA dataset. For a sample $(X, Y) \sim \mathcal{D}_{\text{QA}}$, where $Y = \{y_1, \dots, y_L\}$ represents the answer sequence, the loss is:
$$\mathcal{L}_{\mathrm{LM}} = - \mathbb{E}_{(X, Y) \sim \mathcal{D}_{\text{QA}}} \left[ \sum_{i=1}^{L} \log P(y_i \mid y_{<i}, X; \theta) \right].$$
This stage ensures the model retains its audio-video understanding while adapting to streaming inputs.


\paragraph{Stage 2: Time-Aware Decision Making}
With the backbone adapted to streaming inputs, this stage activates the speak head to learn \textit{when} to respond. We formulate response timing as a binary classification task at each multimodal unit step. The positive labels are task-dependent: for proactive alerts, valid triggers lie within the event window; for narration, they align with segment boundaries. To mitigate trigger sparsity, we balance the loss using $w_{\mathrm{pos}} = N_{\mathrm{neg}} / N_{\mathrm{pos}}$ derived from dataset statistics.

Let $p_t$ be the speak head's predicted probability at time step $t$, and $z_t \in \{0, 1\}$ be the ground truth label. The timing loss is formulated as a weighted Binary Cross-Entropy (BCE):
\begin{multline*}
\mathcal{L}_{\mathrm{time}}
= - \mathbb{E}_{X \sim \mathcal{D}_{\text{stream}}}
\Bigg[\frac{1}{T}\sum_{t=1}^{T}
\Big( w_{\mathrm{pos}}\, z_t \log p_t \\
+ (1-z_t)\log(1-p_t) \Big)\Bigg].
\end{multline*}
To prevent generation quality degradation while optimizing purely for timing, we mix a small portion of the Stage 1 reactive QA data ($\mathcal{D}_{\text{QA}}$) during training. The final objective is a joint optimization:
$$\mathcal{L}_{\text{total}} = \mathcal{L}_{\mathrm{time}} + \lambda \cdot \mathcal{L}_{\mathrm{LM}},$$
where $\mathcal{L}_{\mathrm{LM}}$ is calculated only on the mixed QA samples to maintain linguistic competence, and $\lambda$ balances the two objectives.


\subsubsection{Inference Procedure}
During inference, we strictly follow the training configuration. Video frames are uniformly sampled at 2 fps, and each frame is resized so that the number of pixels does not exceed 65,536. We maintain a persistent KV cache across the stream, so each step only encodes the current multimodal unit. Under this setup, encoding one unit takes 0.3697 seconds on average.
\section{Unified Streaming Evaluation Framework}

\label{sec:eval}
Effective streaming understanding demands models capable of answering queries and autonomously determining interaction timing. Addressing the fragmentation in existing benchmarks (Table~\ref{tab:related_bench}), we establish a unified framework comprising two primary settings: \textbf{proactive interaction}, where the model autonomously monitors the stream to trigger responses, and \textbf{reactive interaction}, where it answers queries based on accumulated context.





\subsection{Proactive Streaming Interaction}
\label{subsec:eval_proactive}
In the proactive setting, the model receives an instruction at the start and must process the stream to determine both the precise timing and content of the response. We categorize this into two sub-tasks: event-driven alert and real-time narration.

\subsubsection{Event-Driven Alert}
This task evaluates the model's temporal awareness, specifically its ability to detect transient events and trigger immediate notifications. We assess this capability under two settings.

\paragraph{Static Temporal Grounding.} 
Following MMDuet on QVHighlights~\cite{lei2021detecting} and Charades-STA~\cite{gao2017tall}, 
\ours incrementally predicts response probabilities for each multimodal unit. 
For QVHighlights, we rank timestamps by normalized probabilities, reporting mAP (ranking quality) and HIT@1 (top-1 accuracy). For localization on Charades-STA, we threshold probabilities to predict spans, reporting R@0.5 and R@0.7 (recall at 0.5 and 0.7 temporal overlap).

\paragraph{Dynamic Streaming Decision.} 
This configuration enforces a strict streaming protocol where the model makes instantaneous decisions conditioned exclusively on the current multimodal unit. We conduct a comprehensive evaluation across OmniMMI (PA), StreamingBench (PO), and OVO-Bench (CRR, REC), spanning both single-event alerts and multi-event recurrence. Specifically, for OVO-Bench, we reformulate the original QA-centric annotations into streaming alert targets to evaluate instantaneous responsiveness. To mitigate transient probability fluctuations, we employ a sliding window mechanism. Success is determined by the temporal inclusion of the autonomously triggered response within the ground-truth interval.

See Appendix~\ref{app:eval_detail} for detailed settings.



\subsubsection{Real-Time Narration}
We define streaming narration as the incremental summarization of evolving events devoid of future context. To evaluate this capability, we employ two settings: a continuous YouCook2 adaptation, constructed by concatenating annotated clips to enforce generation at segment transitions, and the OVO-Bench (SSR) task, where responses are triggered via prediction thresholds and appended to the streaming context. Performance is assessed using the F1 score for temporal localization, BERTScore for the semantic quality of aligned responses, and a GPT-4o-based evaluation of coherence, alignment, and conciseness (detailed in Figure~\ref{fig:prompt-narration}).



\subsection{Reactive QA}
In the reactive setting, the model must interpret temporal evolution to answer questions constrained to the causal video history. 
We utilize OVO-Bench and StreamingBench for standardized evaluation, employing text-based queries to ensure fairness against VideoLLMs baselines and reporting accuracy. To further approximate real-world interaction, we extend the assessment to Video-MME and EgoSchema using synthesized speech inputs. This setting evaluates comprehensive audio-video understanding, with open-ended responses scored by GPT-4o (detailed in Figure~\ref{fig:open-eval-prompt}).

\section{Experiment}
\subsection{Implementation Details}
To address trigger sparsity, we set the positive weight $w_{\mathrm{pos}}=3$ in the weighted BCE loss. For inference, we adopt a pipelined real-time approximation: the model processes unit $t$ while simultaneously acquiring unit $t+1$. To ensure synchronization, we cap generation at 25 tokens (approx. 1s) per segment, allowing longer responses to continue across subsequent units. Please refer to Appendix~\ref{app:impl} for detailed training configurations and complete decoding protocols.


\subsection{Experimental Results}
\paragraph{Baseline Methods}
In the proactive setting, we limit comparison to streaming-capable models: VideoLLM-Online (the basis for many efficiency-focused architectures), MMDuet, and Dispider. We reproduce results using accessible implementations, defaulting to reported figures otherwise. For Reactive QA, we benchmark against representative streaming VideoLLMs. To assess full-modality understanding, we extend evaluation to open-source omni-modal models, including Qwen2.5-Omni, MiniCPM-o, and VITA-1.5.


\begin{table}[h]
\centering
\begin{tblr}{
  width = \linewidth,
  colspec = {X[l] c c},
  colsep = 1.5pt,
  rowsep = 0.8pt,
  stretch = 0.9,
  rows = {font=\normalsize},
  row{1} = {font=\bfseries\normalsize},
  row{2} = {font=\small},
  cell{1}{1} = {r=2}{c, valign=m},
  cells = {valign=m},
  hline{1,3,Z} = {1pt, solid},
  hline{2} = {2-3}{0.6pt, solid},
  hline{9} = {0.6pt, solid, fg=tblRuleLight},
  row{8} = {bg=tblOurs, font=\bfseries\normalsize},
  cell{9,12}{1} = {c=3}{l, font=\itshape\small, bg=tblSection},
}
  Method & QVHighlight & Charades-STA \\
         & mAP / HIT@1 & R@0.5 / 0.7 \\
  TimeChat   & 14.5 / 23.9   & 32.2 / 13.4 \\
  VTimeLLM   & --            & 31.2 / 11.4 \\
  HawkEye    & --            & 31.4 / 14.5 \\
  VTG-LLM    & 16.5 / 33.5   & 33.8 / 15.7 \\
  MMDuet     & 31.3 / 49.6   & 42.4 / 18.0 \\
  Ours       & 53.7 / 53.0   & 44.3 / 19.9 \\
  - Ablation Study & & \\
  Mixed Training & 50.3 / 44.7 & 28.2 / 10.1 \\
  $K=1$          & 46.4 / 47.4 & 32.4 / 13.1 \\
  - Sensitivity Analysis & & \\
  $w_{pos}=2$           & 47.5 / 52.5 & 42.2 / 18.4 \\
  $w_{pos}=4$           & 47.3 / 49.1 & 38.0 / 16.4 \\
\end{tblr}
\caption{Comparison with existing methods on QVHighlights and Charades-STA benchmarks.}
\label{tab:pro_static}
\vspace{-5pt}
\end{table}

\begin{table}[h]
\vspace{-3pt}
\centering
\begin{tblr}{
  width = \linewidth,
  colspec = {X[l,1.5] c c c c},
  colsep = 4.5pt,
  rowsep = 0.8pt,
  stretch = 0.9,
  rows = {font=\normalsize},
  row{1} = {font=\bfseries\normalsize},
  cells = {valign=m},
  hline{1,2} = {1pt, solid},         
  hline{Z}   = {1pt, solid},
  hline{7}   = {0.5pt, fg=tblRuleLight},
  row{6} = {bg=tblOurs, font=\normalsize},
  cell{7,11}{1} = {c=5}{l, font=\itshape\small, bg=tblSection},
}
  Method & PA & PO & CRR & REC \\
  VideoLLM-online & 0.50  & 4.13  & 27.08 & 14.29 \\
  MMDuet          & 22.00 & 29.44 & 16.67 & 12.77 \\
  Dispider        & --    & 25.34 & \textbf{48.75} & 18.05 \\
  M4-a            & 25.50 & --    & --    & --    \\
  Ours            & \textbf{37.50} & \textbf{53.60} & 35.42 & \textbf{33.81} \\
  - Ablation Study & & & & \\
  Mixed Training   & 34.50 & 50.80 & 25.00 & 13.13 \\
  w/o Speak Head   & 12.50 & 12.00 & 0.00  & 6.46  \\
  $K=1$            & 26.00 & 56.40 & 31.25 & 24.32 \\
  - Sensitivity Analysis & & & & \\
  $w_{pos}=2$     & 31.00 & 52.76 & 39.58 & 31.54 \\
  $w_{pos}=4$     & 31.00 & 52.15 & 37.50 & 26.74 \\
\end{tblr}
\caption{Comparison across single-alert (PA, PO, CRR) and recurring alert (REC) benchmarks.}
\label{tab:pro_dy}
\vspace{-10pt}
\end{table}

\paragraph{Event-Driven Alert}
In static temporal grounding (Table~\ref{tab:pro_static}), \ours advances temporal localization on QVHighlights (53.7 mAP) and Charades-STA (44.3/19.9 R@0.5/0.7), confirming that incremental speak probabilities provide enhanced temporal saliency for precise ranking and prediction. 
In the dynamic setting (Table~\ref{tab:pro_dy}), \ours demonstrates strong efficacy on single-alert tasks: it excels on PA and PO while remaining competitive on CRR, validating its precise proactive triggering and robust evidence accumulation. Furthermore, \ours dominates on the REC benchmark, validating its recurrence modeling for tracking repeated instances.

\paragraph{Real-Time Narration}
As shown in Table~\ref{tab:narration}, \ours achieves the best temporal triggering accuracy, obtaining an F1 score of 35.21 on YouCook2 and 14.54 on OVO-Bench (SSR), which indicates more precise alignment between generated responses and the annotated narration windows. 
It also achieves the highest GPT-4o score on both benchmarks. This score averages three criteria (story coherence, alignment to ground truth, and conciseness), with the per-criterion breakdown in Table~\ref{tab:narration_full}, suggesting more coherent and better-aligned narration when generation is triggered online and the outputs are carried forward as context.
\begin{table}[h]
\vspace{-6pt}
\centering
\begin{tblr}{
  width = \linewidth,
  colspec = {X[l,2.2] X[c] X[c] X[c] X[c] X[c] X[c]}, 
  colsep = 1pt,            
  rowsep = 0.78pt,            
  stretch = 0.9,           
  rows = {font=\small},    
  row{1} = {font=\bfseries\small},
  row{2} = {font=\footnotesize}, 
  cell{1}{1} = {r=2}{c, valign=m},           
  cell{1}{2,5} = {c=3}{c, font=\bfseries},   
  cells = {valign=m},
  vline{5} = {0.6pt, fg=black!35}, 
  hline{1,3} = {1pt, solid},
  hline{2} = {2-7}{0.5pt, solid},
  hline{Z} = {1pt, solid},
  hline{8} = {0.5pt, fg=tblRuleLight}, 
  row{7} = {bg=tblOurs, font=\small}, 
  cell{8,12}{1} = {c=4}{l, font=\itshape\footnotesize, bg=tblSection}, 
  cell{8,12}{5} = {c=3}{bg=tblSection}, 
}
  Method & YouCook2 & & & OVO-Bench (SSR)  & & \\
  & F1 & BERT & GPT & F1 & BERT & GPT \\
  
  TimeChat        & 21.70 & --   & --             & --    & --   & -- \\
  VTG-LLM         & 17.50 & --   & --             & --    & --   & -- \\
  VideoLLM-online & 18.82 & 0.82 & 0.17 & 10.24 & \textbf{0.84} & 0.18 \\
  MMDuet          & 17.81 & 0.83 & 0.23 & 9.02  & 0.79 & 0.31 \\
  
  \textbf{Ours}            & \textbf{35.21} & \textbf{0.83} & \textbf{0.39} & \textbf{14.54} & 0.83 & \textbf{0.42} \\
  - Ablation Study & & & & & & \\
  Mixed Training  & 31.42 & 0.81 & 0.34 & 8.88  & 0.80 & 0.33 \\
  w/o Speak Head   & 9.25    & 0.79   & 0.24            & 3.39    & 0.77   & 0.26 \\
  $K=1$           & 34.43    & 0.82   & 0.37             & 9.64    & 0.78   & 0.32 \\
  
  - Sensitivity Analysis & & & & & & \\
  $w_{pos}=2$     & 27.82 & 0.83 & 0.45 & 10.38 & 0.57 & 0.38 \\
  $w_{pos}=4$     & 35.55 & 0.81 & 0.47 & 13.48 & 0.75 & 0.34 \\
\end{tblr}
\caption{Streaming narration results on YouCook2 and OVO-Bench (SSR). We report F1 for temporal window alignment, and use BERTScore and averaged GPT-4o scores to assess narration quality.}
\label{tab:narration}
\vspace{-10pt}
\end{table}

\paragraph{Reactive QA}
On OVO-Bench (Table~\ref{tab:ovobench}), \ours leads in both ``Real-time Visual Perception'' and ``Backward Tracing''. Its superiority over streaming baselines highlights enhanced sensitivity to time-localized cues and robust utilization of historical evidence under truncated contexts. On StreamingBench (Table~\ref{tab:streamingbench}), \ours maintains high accuracy and secures the top rank on ``Omni-Source Understanding'' benchmark. attributed to preserving aligned audio during training, which bolsters audio–visual integration. In full-modality evaluation (Table~\ref{tab:turn}), \ours attains the best performance on Video-MME (without subtitles) and remains competitive on EgoSchema. Notably, these results utilize spoken queries with joint audio–visual inputs to approximate conversational interaction, distinct from text-prompted prior work.

Overall, \ours strengthens temporal awareness and streaming decision-making, optimizing timing and content via audio–video joint modeling.
\begin{table}[htbp]
\vspace{-2pt}
\centering
\begin{tblr}{
  width = \linewidth,
  colspec = {X[l] c c}, 
  colsep = 4pt,
  rowsep = 0.8pt,
  stretch = 0.9,
  rows = {font=\normalsize},
  row{1} = {font=\bfseries\normalsize},
  cells = {valign=m},
  hline{1,Z} = {1pt, solid},   
  hline{2} = {0.5pt, solid},   
  hline{6} = {0.5pt, fg=tblRuleLight}, 
  row{5} = {bg=tblOurs, font=\normalsize}, 
  cell{6,10}{1} = {c=3}{l, font=\itshape\small, bg=tblSection},
}
  Method & Video-MME & EgoSchema \\
  
  Qwen2.5-Omni    & 20.50 & \textbf{58.40} \\
  VITA-1.5        & 28.56 & 45.40 \\
  MiniCPM-o       & 19.37 & 55.20 \\
  
  \textbf{Ours}            & \textbf{33.30} & 55.40 \\
  
  - Ablation Study & & \\ 
  Mixed Training  & 33.00 & 50.20 \\
  w/o speak head   & 9.11    & 12.80   \\
  $K=1$ & 34.56 & 54.00 \\
  
  - Sensitivity Analysis & & \\
  $w_{pos}=2$     & 33.20 & 52.60 \\
  $w_{pos}=4$     & 33.10 & 54.80 \\
\end{tblr}
\caption{Full-modality QA results on Video-MME (no subtitles) and EgoSchema, evaluated with spoken questions to approximate real conversational interaction.}
\label{tab:turn}
\vspace{-10pt}
\end{table}
\begin{table*}[t]
\centering
\begin{tblr}{
  width = \linewidth,
  colspec = {X[l,2.25] *{9}{X[c,1]}}, 
  colsep = 1pt,
  rowsep = 0.78pt,
  stretch = 0.87,
  cells = {valign=m},
  rows = {font=\normalsize},
  column{1} = {font=\normalsize},
  row{1} = {font=\bfseries\normalsize, halign=c},
  row{2} = {font=\small, halign=c},
  cell{1}{1}  = {r=2}{c, valign=m, font=\normalsize},
  cell{1}{2}  = {c=6}{c, font=\normalsize},
  cell{1}{8}  = {c=3}{c, font=\normalsize},
  vline{8} = {0.5pt, fg=black!35}, 
  hline{1} = {1pt},
  hline{2} = {2-10}{0.5pt}, 
  hline{3} = {1pt},
  hline{8} = {0.5pt, fg=tblRuleLight},
  hline{Z} = {1pt},
  row{7} = {bg=tblOurs, font=\normalsize},
  cell{9-11}{1} = {font=\normalsize},
  row{8,12} = {bg=tblSection, font=\itshape\small},
  cell{8,12}{1} = {l},
  cell{8,12}{2} = {c=6}{},
  cell{8,12}{8} = {c=3}{}
}
Method & Real-time Visual Perception & & & & & & Backward Tracing & & \\
& OCR & ACR & ATR & STU & FPD & OJR & EPM & ASI & HLD \\
VideoLLM-online  & 8.05  & 23.85 & 12.07 & 14.04 & 45.54 & 21.20 & 22.22 & 18.80 & 12.18 \\
MMDuet           & 13.42 & 11.93 & 14.66 & 11.80 & 14.85 & 10.33 & 10.44 & 8.78  & 0.54  \\
Dispider         & 57.72 & 49.54 & 62.07 & \textbf{44.94} & 61.39 & 51.63 & 48.48 & \textbf{55.41} & 4.30  \\
Flash-VStream-7B & 24.16 & 29.36 & 28.45 & 33.71 & 25.74 & 28.80 & 39.06 & 37.16 & 5.91  \\
\textbf{Ours}    & \textbf{63.09} & \textbf{53.21} & \textbf{68.10} & 39.33 & \textbf{69.31} & \textbf{58.15} & \textbf{55.89} & 47.30 & \textbf{23.66} \\

- Ablation Study & & & & & & & & & \\
Mixed Training   & 63.09 & 55.05 & 63.79 & 37.64 & 61.39 & 55.43 & 55.22 & 45.95 & 27.96 \\
w/o Speak Head    & 61.07 & 55.05 & 63.97 & 39.89 & 65.35 & 54.89 & 53.87 & 47.97 & 29.03 \\
$K=1$            & 61.47 & 55.05 & 68.10 & 39.89 & 65.35 & 60.33 & 56.57 & 46.62 & 20.97 \\

- Sensitivity Analysis & & & & & & & & & \\
$w_{pos}=2$      & 64.43 & 51.38 & 68.97 & 39.33 & 64.36 & 60.87 & 54.88 & 46.62 & 20.97 \\
$w_{pos}=4$      & 65.10 & 54.13 & 68.97 & 38.20 & 70.30 & 61.41 & 56.57 & 46.27 & 22.58 \\
\end{tblr}
\caption{Reactive QA results on OVO-Bench (excluding Forward Active Responding), evaluating time-sensitive understanding across Real-time Visual Perception and Backward Tracing.}
\label{tab:ovobench}
\end{table*}

\begin{table*}[h!]
\vspace{-4pt}
\centering
\begin{tblr}{
  width = \linewidth,
  colspec = {X[l,3.6] *{17}{X[c,1]}},
  colsep = 1.2pt, 
  rowsep = 0.88pt,
  stretch = 0.9,
  cells = {valign=m},
  rows = {font=\small},         
  row{1} = {font=\small, halign=c}, 
  row{2} = {font=\small, halign=c},   
  column{1} = {font=\small}, 
  cell{1}{1}  = {r=2}{c, valign=m},
  cell{1}{2}  = {c=10}{c}, 
  cell{1}{12} = {c=4}{c},  
  cell{1}{16} = {c=3}{c},  
  vline{12,16} = {0.5pt, fg=black!35},
  hline{1,3,Z} = {1pt},
  hline{2} = {2-18}{0.5pt}, 
  hline{7} = {0.5pt, fg=tblRuleLight},
  row{6} = {bg=tblOurs}, 
  row{7,11} = {bg=tblSection, font=\itshape\small},
  cell{7,11}{1} = {l},     
  cell{7,11}{2} = {c=10}{}, 
  cell{7,11}{12} = {c=4}{}, 
  cell{7,11}{16} = {c=3}{}, 
}
  Method & {Real-Time Visual Understanding} & & & & & & & & & & {Omni-Source \\ Understanding} & & & & {Contextual \\ Understanding} & & \\
  & OP & CR & CS & ATP & EU & TR & PR & SU & ACP & CT & ER & SCU & SD & MA & ACU & MCU & SQA \\
  VideoLLM-Online & 39.07 & 40.06 & 34.49 & 31.05 & 45.96 & 32.40 & 31.48 & 34.16 & 42.49 & 27.89 & 31.20 & 26.51 & 24.10 & 32.00 & 24.19 & 29.20 & 26.55 \\
  Flash-VStream   & 25.89 & 43.57 & 24.91 & 23.87 & 27.33 & 13.08 & 18.52 & 25.20 & 23.87 & 48.70 & 25.91 & 24.90 & 25.60 & 28.40 & 24.80 & 25.20 & 24.12 \\
  Dispider        & 74.92 & 75.53 & 74.10 & 73.08 & 74.44 & 59.52 & 76.14 & \textbf{62.91} & 62.16 & 45.80 & 35.46 & 25.26 & 38.57 & 43.34 & \textbf{39.62} & 27.65 & 33.61 \\
  \textbf{Ours}            & \textbf{76.96} & \textbf{78.91} & \textbf{77.92} & \textbf{82.05} & \textbf{74.84} & \textbf{72.90} & \textbf{82.41} & 61.79 & \textbf{65.91} & \textbf{51.06} & \textbf{40.40} & \textbf{34.80} & \textbf{50.40} & \textbf{58.80} & 37.60 & \textbf{34.00} & \textbf{44.47} \\
  - Ablation Study  & & & & & & & & & & & & & & & & & \\
  Mixed Training  & 75.51 & 85.71 & 76.19 & 78.23 & 59.77 & 61.05 & 73.21 & 60.00 & 59.67 & 23.38 & 38.80 & 26.00 & 40.80 & 47.20 & 35.60 & 27.20 & 26.80 \\
  w/o Speak Head   & 76.13    & 70.49    & 74.14    & 82.40    & 72.86    & 70.80    & 84.78    & 63.20    & 64.91    & 51.69    & 39.75    & 30.36    & 45.87    & 47.20    & 35.27    & 24.79    & 24.80    \\
  $K=1$           & 76.69 & 82.03 & 78.86 & 82.05 & 74.84 & 72.90 & 79.63 & 59.76 & 64.49 & 50.53 & 38.80 & 29.60 & 46.40 & 51.60 & 33.20 & 27.60 & 22.80 \\
  - Sensitivity Analysis & & & & & & & & & & & & & & & & & \\
  $w_{pos}=2$     & 75.61 & 81.25 & 76.97 & 82.37 & 71.70 & 75.08 & 81.48 & 62.20 & 65.62 & 50.00 & 39.60 & 30.80 & 59.38 & 52.40 & 35.60 & 28.00 & 26.40 \\
  $w_{pos}=4$     & 75.61 & 80.47 & 79.18 & 82.37 & 73.58 & 75.08 & 82.41 & 63.10 & 65.34 & 47.34 & 39.60 & 28.80 & 44.40 & 51.60 & 35.60 & 28.40 & 26.40 \\
\end{tblr}
\caption{Reactive QA results on StreamingBench (excluding PO), evaluating real-time understanding under streaming input across Real-Time Visual Understanding, Omni-Source Understanding, and Contextual Understanding.}
\label{tab:streamingbench}
\vspace{-8pt}
\end{table*}

\subsection{Ablation Study}
\paragraph{Single-Stage vs. Two-Stage Training}
We validate the two-stage curriculum by mixing all data and training directly with the stage-2 objective. This variant consistently degrades on tasks that require online timing and triggering, most notably on dynamic decision making (e.g., REC) and streaming narration (Table~\ref{tab:pro_dy}, Table~\ref{tab:narration}). The results indicate that progressive training is important for learning well-calibrated temporal decision making under streaming input.

\paragraph{Speak Head for Response Gating}
We replace the speak head with a `<|silence|>' token following prior work, and cast triggering as next-token prediction with a reweighted loss. Lacking explicit probabilities, we omit QVHighlights and Charades-STA, instead evaluating triggering based on the first non-`<|silence|>' token.


\paragraph{Last-Layer vs. Last-4-Layer Aggregation}
We ablate four-layer aggregation by restricting the speak head to the final layer ($K{=}1$). This notably degrades temporal grounding and dynamic triggering (Tables~\ref{tab:pro_static}, \ref{tab:pro_dy}) while leaving timestamp-conditioned understanding largely unaffected (Tables~\ref{tab:ovobench}, \ref{tab:streamingbench}). This confirms multi-layer aggregation yields robust signals essential for streaming.

\subsection{Sensitivity analysis}

We sweep the positive weight $w_{\mathrm{pos}}$ in the weighted BCE loss of the speak head to mitigate the class imbalance from sparse speaking timestamps. We observe that $w_{\mathrm{pos}}$ is critical for proactive tasks (Tables~\ref{tab:pro_static}--\ref{tab:narration}), while reactive understanding and full-modality QA remain insensitive (Tables~\ref{tab:turn}, \ref{tab:ovobench}). Overall, $w_{\mathrm{pos}}=3$ yields the most balanced performance. See Appendix~\ref{app:sen} for sensitivity analysis on inference-time triggering thresholds.

\section{Conclusion}

We introduce \ours, a real-time omni-multimodal assistant that redefines streaming interaction as the unification of proactive and reactive paradigms. \ours is the first framework to excel in both modes. To achieve this, we construct a streaming dataset and training recipe that enhance temporal modeling and decision-making. Furthermore, we standardize evaluation through a unified protocol tailored to this dual paradigm, where \ours demonstrates superior performance. Finally, we provide a systematized overview of prior methods to facilitate future research.



\section*{Limitations}
While optimized for streaming interaction, the model remains susceptible to distortions such as signal degradation and audio–video asynchrony. Additionally, while capable of continuous streaming, capturing extremely long-term dependencies spanning hours remains constrained by finite context windows and memory. Finally, optimizing the trade-off between inference efficiency and response quality under strict resource constraints remains a critical direction for future work.

\section*{Ethical Statement}
This work utilizes publicly available datasets consistent with their original licenses. While ROMA enables proactive monitoring capabilities, we acknowledge the potential risk of misuse for unauthorized surveillance or privacy infringement. This model is intended for research purposes; due to the possibility of hallucinations or biases inherited from the base LLM, human oversight is strictly required for critical real-world applications.


\bibliography{custom}

@article{defossez2024moshi,
  title={Moshi: a speech-text foundation model for real-time dialogue},
  author={D{\'e}fossez, Alexandre and Mazar{\'e}, Laurent and Orsini, Manu and Royer, Am{\'e}lie and P{\'e}rez, Patrick and J{\'e}gou, Herv{\'e} and Grave, Edouard and Zeghidour, Neil},
  journal={arXiv preprint arXiv:2410.00037},
  year={2024}
}

@article{shi2025voila,
  title={Voila: Voice-Language Foundation Models for Real-Time Autonomous Interaction and Voice Role-Play},
  author={Shi, Yemin and Shu, Yu and Dong, Siwei and Liu, Guangyi and Sesay, Jaward and Li, Jingwen and Hu, Zhiting},
  journal={arXiv preprint arXiv:2505.02707},
  year={2025}
}

@article{qian2024streaming,
  title={Streaming long video understanding with large language models},
  author={Qian, Rui and Dong, Xiaoyi and Zhang, Pan and Zang, Yuhang and Ding, Shuangrui and Lin, Dahua and Wang, Jiaqi},
  journal={Advances in Neural Information Processing Systems},
  volume={37},
  pages={119336--119360},
  year={2024}
}

@article{zhang2024flash,
  title={Flash-vstream: Memory-based real-time understanding for long video streams},
  author={Zhang, Haoji and Wang, Yiqin and Tang, Yansong and Liu, Yong and Feng, Jiashi and Dai, Jifeng and Jin, Xiaojie},
  journal={arXiv preprint arXiv:2406.08085},
  year={2024}
}

@article{wang2024videollamb,
  title={Videollamb: Long-context video understanding with recurrent memory bridges},
  author={Wang, Yuxuan and Xie, Cihang and Liu, Yang and Zheng, Zilong},
  journal={arXiv preprint arXiv:2409.01071},
  year={2024}
}

@article{zhang2024internlm,
  title={Internlm-xcomposer2. 5-omnilive: A comprehensive multimodal system for long-term streaming video and audio interactions},
  author={Zhang, Pan and Dong, Xiaoyi and Cao, Yuhang and Zang, Yuhang and Qian, Rui and Wei, Xilin and Chen, Lin and Li, Yifei and Niu, Junbo and Ding, Shuangrui and others},
  journal={arXiv preprint arXiv:2412.09596},
  year={2024}
}

@article{xiong2025streaming,
  title={Streaming video understanding and multi-round interaction with memory-enhanced knowledge},
  author={Xiong, Haomiao and Yang, Zongxin and Yu, Jiazuo and Zhuge, Yunzhi and Zhang, Lu and Zhu, Jiawen and Lu, Huchuan},
  journal={arXiv preprint arXiv:2501.13468},
  year={2025}
}

@article{wang2025streambridge,
  title={StreamBridge: Turning Your Offline Video Large Language Model into a Proactive Streaming Assistant},
  author={Wang, Haibo and Feng, Bo and Lai, Zhengfeng and Xu, Mingze and Li, Shiyu and Ge, Weifeng and Dehghan, Afshin and Cao, Meng and Huang, Ping},
  journal={arXiv preprint arXiv:2505.05467},
  year={2025}
}

@article{zhao2025cogstream,
  title={CogStream: Context-guided Streaming Video Question Answering},
  author={Zhao, Zicheng and Wang, Kangyu and Li, Shijie and Qian, Rui and Lin, Weiyao and Liu, Huabin},
  journal={arXiv preprint arXiv:2506.10516},
  year={2025}
}

@article{di2025streaming,
  title={Streaming video question-answering with in-context video kv-cache retrieval},
  author={Di, Shangzhe and Yu, Zhelun and Zhang, Guanghao and Li, Haoyuan and Zhong, Tao and Cheng, Hao and Li, Bolin and He, Wanggui and Shu, Fangxun and Jiang, Hao},
  journal={arXiv preprint arXiv:2503.00540},
  year={2025}
}

@article{ning2025livevlm,
  title={LiveVLM: Efficient Online Video Understanding via Streaming-Oriented KV Cache and Retrieval},
  author={Ning, Zhenyu and Liu, Guangda and Jin, Qihao and Ding, Wenchao and Guo, Minyi and Zhao, Jieru},
  journal={arXiv preprint arXiv:2505.15269},
  year={2025}
}

@article{yang2025streamagent,
  title={Streamagent: Towards anticipatory agents for streaming video understanding},
  author={Yang, Haolin and Tang, Feilong and Zhao, Lingxiao and An, Xiang and Hu, Ming and Li, Huifa and Zhuang, Xinlin and Lu, Yifan and Zhang, Xiaofeng and Swikir, Abdalla and others},
  journal={arXiv preprint arXiv:2508.01875},
  year={2025}
}

@article{xu2025streamingvlm,
  title={Streamingvlm: Real-time understanding for infinite video streams},
  author={Xu, Ruyi and Xiao, Guangxuan and Chen, Yukang and He, Liuning and Peng, Kelly and Lu, Yao and Han, Song},
  journal={arXiv preprint arXiv:2510.09608},
  year={2025}
}

@article{chen2025streamkv,
  title={StreamKV: Streaming Video Question-Answering with Segment-based KV Cache Retrieval and Compression},
  author={Chen, Yilong and Bai, Xiang and Wang, Zhibin and Bai, Chengyu and Dai, Yuhan and Lu, Ming and Zhang, Shanghang},
  journal={arXiv preprint arXiv:2511.07278},
  year={2025}
}

@inproceedings{chen2025livecc,
  title={Livecc: Learning video llm with streaming speech transcription at scale},
  author={Chen, Joya and Zeng, Ziyun and Lin, Yiqi and Li, Wei and Ma, Zejun and Shou, Mike Zheng},
  booktitle={Proceedings of the Computer Vision and Pattern Recognition Conference},
  pages={29083--29095},
  year={2025}
}

@inproceedings{chen2024videollm,
  title={Videollm-online: Online video large language model for streaming video},
  author={Chen, Joya and Lv, Zhaoyang and Wu, Shiwei and Lin, Kevin Qinghong and Song, Chenan and Gao, Difei and Liu, Jia-Wei and Gao, Ziteng and Mao, Dongxing and Shou, Mike Zheng},
  booktitle={Proceedings of the IEEE/CVF Conference on Computer Vision and Pattern Recognition},
  pages={18407--18418},
  year={2024}
}

@article{yang2025assistpda,
  title={AssistPDA: An Online Video Surveillance Assistant for Video Anomaly Prediction, Detection, and Analysis},
  author={Yang, Zhiwei and Gao, Chen and Liu, Jing and Wu, Peng and Pang, Guansong and Shou, Mike Zheng},
  journal={arXiv preprint arXiv:2503.21904},
  year={2025}
}

@inproceedings{li2025lion,
  title={Lion-fs: Fast \& slow video-language thinker as online video assistant},
  author={Li, Wei and Hu, Bing and Shao, Rui and Shen, Leyang and Nie, Liqiang},
  booktitle={Proceedings of the Computer Vision and Pattern Recognition Conference},
  pages={3240--3251},
  year={2025}
}

@article{wu2024videollm,
  title={Videollm-mod: Efficient video-language streaming with mixture-of-depths vision computation},
  author={Wu, Shiwei and Chen, Joya and Lin, Kevin Qinghong and Wang, Qimeng and Gao, Yan and Xu, Qianli and Xu, Tong and Hu, Yao and Chen, Enhong and Shou, Mike Zheng},
  journal={Advances in Neural Information Processing Systems},
  volume={37},
  pages={109922--109947},
  year={2024}
}

@article{yang2025livestar,
  title={LiveStar: Live Streaming Assistant for Real-World Online Video Understanding},
  author={Yang, Zhenyu and Zhang, Kairui and Hu, Yuhang and Wang, Bing and Qian, Shengsheng and Wen, Bin and Yang, Fan and Gao, Tingting and Dong, Weiming and Xu, Changsheng},
  journal={arXiv preprint arXiv:2511.05299},
  year={2025}
}

@article{wang2024videollm,
  title={Videollm knows when to speak: Enhancing time-sensitive video comprehension with video-text duet interaction format},
  author={Wang, Yueqian and Meng, Xiaojun and Wang, Yuxuan and Liang, Jianxin and Wei, Jiansheng and Zhang, Huishuai and Zhao, Dongyan},
  journal={arXiv preprint arXiv:2411.17991},
  year={2024}
}

@inproceedings{qian2025dispider,
  title={Dispider: Enabling video llms with active real-time interaction via disentangled perception, decision, and reaction},
  author={Qian, Rui and Ding, Shuangrui and Dong, Xiaoyi and Zhang, Pan and Zang, Yuhang and Cao, Yuhang and Lin, Dahua and Wang, Jiaqi},
  booktitle={Proceedings of the Computer Vision and Pattern Recognition Conference},
  pages={24045--24055},
  year={2025}
}

@article{yao2024minicpm,
  title={Minicpm-v: A gpt-4v level mllm on your phone},
  author={Yao, Yuan and Yu, Tianyu and Zhang, Ao and Wang, Chongyi and Cui, Junbo and Zhu, Hongji and Cai, Tianchi and Li, Haoyu and Zhao, Weilin and He, Zhihui and others},
  journal={arXiv preprint arXiv:2408.01800},
  year={2024}
}

@article{xu2025qwen2,
  title={Qwen2. 5-omni technical report},
  author={Xu, Jin and Guo, Zhifang and He, Jinzheng and Hu, Hangrui and He, Ting and Bai, Shuai and Chen, Keqin and Wang, Jialin and Fan, Yang and Dang, Kai and others},
  journal={arXiv preprint arXiv:2503.20215},
  year={2025}
}

@article{Qwen3-Omni,
  title={Qwen3-Omni Technical Report},
  author={Jin Xu and Zhifang Guo and Hangrui Hu and Yunfei Chu and Xiong Wang and Jinzheng He and Yuxuan Wang and Xian Shi and Ting He and Xinfa Zhu and Yuanjun Lv and Yongqi Wang and Dake Guo and He Wang and Linhan Ma and Pei Zhang and Xinyu Zhang and Hongkun Hao and Zishan Guo and Baosong Yang and Bin Zhang and Ziyang Ma and Xipin Wei and Shuai Bai and Keqin Chen and Xuejing Liu and Peng Wang and Mingkun Yang and Dayiheng Liu and Xingzhang Ren and Bo Zheng and Rui Men and Fan Zhou and Bowen Yu and Jianxin Yang and Le Yu and Jingren Zhou and Junyang Lin},
  journal={arXiv preprint arXiv:2509.17765},
  year={2025}
}

@article{zhang2025stream,
  title={Stream-Omni: Simultaneous Multimodal Interactions with Large Language-Vision-Speech Model},
  author={Zhang, Shaolei and Guo, Shoutao and Fang, Qingkai and Zhou, Yan and Feng, Yang},
  journal={arXiv preprint arXiv:2506.13642},
  year={2025}
}

@article{fu2025vispeak,
  title={ViSpeak: Visual Instruction Feedback in Streaming Videos},
  author={Fu, Shenghao and Yang, Qize and Li, Yuan-Ming and Peng, Yi-Xing and Lin, Kun-Yu and Wei, Xihan and Hu, Jian-Fang and Xie, Xiaohua and Zheng, Wei-Shi},
  journal={arXiv preprint arXiv:2503.12769},
  year={2025}
}

@article{lin2024streamingbench,
  title={Streamingbench: Assessing the gap for mllms to achieve streaming video understanding},
  author={Lin, Junming and Fang, Zheng and Chen, Chi and Wan, Zihao and Luo, Fuwen and Li, Peng and Liu, Yang and Sun, Maosong},
  journal={arXiv preprint arXiv:2411.03628},
  year={2024}
}

@inproceedings{niu2025ovo,
  title={OVO-Bench: How Far is Your Video-LLMs from Real-World Online Video Understanding?},
  author={Niu, Junbo and Li, Yifei and Miao, Ziyang and Ge, Chunjiang and Zhou, Yuanhang and He, Qihao and Dong, Xiaoyi and Duan, Haodong and Ding, Shuangrui and Qian, Rui and others},
  booktitle={Proceedings of the Computer Vision and Pattern Recognition Conference},
  pages={18902--18913},
  year={2025}
}

@article{wu2024streambench,
  title={Streambench: Towards benchmarking continuous improvement of language agents},
  author={Wu, Cheng-Kuang and Tam, Zhi Rui and Lin, Chieh-Yen and Chen, Yun-Nung Vivian and Lee, Hung-yi},
  journal={Advances in Neural Information Processing Systems},
  volume={37},
  pages={107039--107063},
  year={2024}
}

@inproceedings{wang2025omnimmi,
  title={OmniMMI: A Comprehensive Multi-modal Interaction Benchmark in Streaming Video Contexts},
  author={Wang, Yuxuan and Wang, Yueqian and Chen, Bo and Wu, Tong and Zhao, Dongyan and Zheng, Zilong},
  booktitle={Proceedings of the Computer Vision and Pattern Recognition Conference},
  pages={18925--18935},
  year={2025}
}

@inproceedings{huang2025online,
  title={Online Video Understanding: OVBench and VideoChat-Online},
  author={Huang, Zhenpeng and Li, Xinhao and Li, Jiaqi and Wang, Jing and Zeng, Xiangyu and Liang, Cheng and Wu, Tao and Chen, Xi and Li, Liang and Wang, Limin},
  booktitle={Proceedings of the Computer Vision and Pattern Recognition Conference},
  pages={3328--3338},
  year={2025}
}

@article{yang2025svbench,
  title={Svbench: A benchmark with temporal multi-turn dialogues for streaming video understanding},
  author={Yang, Zhenyu and Hu, Yuhang and Du, Zemin and Xue, Dizhan and Qian, Shengsheng and Wu, Jiahong and Yang, Fan and Dong, Weiming and Xu, Changsheng},
  journal={arXiv preprint arXiv:2502.10810},
  year={2025}
}

@inproceedings{anne2017localizing,
  title={Localizing moments in video with natural language},
  author={Anne Hendricks, Lisa and Wang, Oliver and Shechtman, Eli and Sivic, Josef and Darrell, Trevor and Russell, Bryan},
  booktitle={Proceedings of the IEEE international conference on computer vision},
  pages={5803--5812},
  year={2017}
}

@inproceedings{epstein2020oops,
  title={Oops! predicting unintentional action in video},
  author={Epstein, Dave and Chen, Boyuan and Vondrick, Carl},
  booktitle={Proceedings of the IEEE/CVF conference on computer vision and pattern recognition},
  pages={919--929},
  year={2020}
}

@inproceedings{gao2017tall,
  title={Tall: Temporal activity localization via language query},
  author={Gao, Jiyang and Sun, Chen and Yang, Zhenheng and Nevatia, Ram},
  booktitle={Proceedings of the IEEE international conference on computer vision},
  pages={5267--5275},
  year={2017}
}

@inproceedings{tang2019coin,
  title={Coin: A large-scale dataset for comprehensive instructional video analysis},
  author={Tang, Yansong and Ding, Dajun and Rao, Yongming and Zheng, Yu and Zhang, Danyang and Zhao, Lili and Lu, Jiwen and Zhou, Jie},
  booktitle={Proceedings of the IEEE/CVF Conference on Computer Vision and Pattern Recognition},
  pages={1207--1216},
  year={2019}
}

@inproceedings{zhou2018towards,
  title={Towards automatic learning of procedures from web instructional videos},
  author={Zhou, Luowei and Xu, Chenliang and Corso, Jason},
  booktitle={Proceedings of the AAAI conference on artificial intelligence},
  volume={32},
  number={1},
  year={2018}
}

@inproceedings{caba2015activitynet,
  title={Activitynet: A large-scale video benchmark for human activity understanding},
  author={Caba Heilbron, Fabian and Escorcia, Victor and Ghanem, Bernard and Carlos Niebles, Juan},
  booktitle={Proceedings of the ieee conference on computer vision and pattern recognition},
  pages={961--970},
  year={2015}
}

@article{wang2023internvid,
  title={Internvid: A large-scale video-text dataset for multimodal understanding and generation},
  author={Wang, Yi and He, Yinan and Li, Yizhuo and Li, Kunchang and Yu, Jiashuo and Ma, Xin and Li, Xinhao and Chen, Guo and Chen, Xinyuan and Wang, Yaohui and others},
  journal={arXiv preprint arXiv:2307.06942},
  year={2023}
}

@article{chen2023egoplan,
  title={Egoplan-bench: Benchmarking multimodal large language models for human-level planning},
  author={Chen, Yi and Ge, Yuying and Ge, Yixiao and Ding, Mingyu and Li, Bohao and Wang, Rui and Xu, Ruifeng and Shan, Ying and Liu, Xihui},
  journal={arXiv preprint arXiv:2312.06722},
  year={2023}
}

@inproceedings{yang2022avqa,
  title={Avqa: A dataset for audio-visual question answering on videos},
  author={Yang, Pinci and Wang, Xin and Duan, Xuguang and Chen, Hong and Hou, Runze and Jin, Cong and Zhu, Wenwu},
  booktitle={Proceedings of the 30th ACM international conference on multimedia},
  pages={3480--3491},
  year={2022}
}

@inproceedings{yao2025timechat,
  title={Timechat-online: 80\% visual tokens are naturally redundant in streaming videos},
  author={Yao, Linli and Li, Yicheng and Wei, Yuancheng and Li, Lei and Ren, Shuhuai and Liu, Yuanxin and Ouyang, Kun and Wang, Lean and Li, Shicheng and Li, Sida and others},
  booktitle={Proceedings of the 33rd ACM International Conference on Multimedia},
  pages={10807--10816},
  year={2025}
}

@article{tenney2019bert,
  title={BERT rediscovers the classical NLP pipeline},
  author={Tenney, Ian and Das, Dipanjan and Pavlick, Ellie},
  journal={arXiv preprint arXiv:1905.05950},
  year={2019}
}

@article{belrose2023eliciting,
  title={Eliciting latent predictions from transformers with the tuned lens},
  author={Belrose, Nora and Furman, Zach and Smith, Logan and Halawi, Danny and Ostrovsky, Igor and McKinney, Lev and Biderman, Stella and Steinhardt, Jacob},
  journal={arXiv preprint arXiv:2303.08112},
  year={2023}
}

@article{seed2025seed1,
  title={Seed1. 5-thinking: Advancing superb reasoning models with reinforcement learning},
  author={Seed, ByteDance and Chen, Jiaze and Fan, Tiantian and Liu, Xin and Liu, Lingjun and Lin, Zhiqi and Wang, Mingxuan and Wang, Chengyi and Wei, Xiangpeng and Xu, Wenyuan and others},
  journal={arXiv preprint arXiv:2504.13914},
  year={2025}
}

@inproceedings{zheng2024llamafactory,
  title={LlamaFactory: Unified Efficient Fine-Tuning of 100+ Language Models},
  author={Yaowei Zheng and Richong Zhang and Junhao Zhang and Yanhan Ye and Zheyan Luo and Zhangchi Feng and Yongqiang Ma},
  booktitle={Proceedings of the 62nd Annual Meeting of the Association for Computational Linguistics (Volume 3: System Demonstrations)},
  address={Bangkok, Thailand},
  publisher={Association for Computational Linguistics},
  year={2024},
  url={http://arxiv.org/abs/2403.13372}
}

@article{zheng2023judging,
  title={Judging llm-as-a-judge with mt-bench and chatbot arena},
  author={Zheng, Lianmin and Chiang, Wei-Lin and Sheng, Ying and Zhuang, Siyuan and Wu, Zhanghao and Zhuang, Yonghao and Lin, Zi and Li, Zhuohan and Li, Dacheng and Xing, Eric and others},
  journal={Advances in neural information processing systems},
  volume={36},
  pages={46595--46623},
  year={2023}
}

@article{hurst2024gpt,
  title={Gpt-4o system card},
  author={Hurst, Aaron and Lerer, Adam and Goucher, Adam P and Perelman, Adam and Ramesh, Aditya and Clark, Aidan and Ostrow, AJ and Welihinda, Akila and Hayes, Alan and Radford, Alec and others},
  journal={arXiv preprint arXiv:2410.21276},
  year={2024}
}

@inproceedings{horvitz1999principles,
  title={Principles of mixed-initiative user interfaces},
  author={Horvitz, Eric},
  booktitle={Proceedings of the SIGCHI conference on Human Factors in Computing Systems},
  pages={159--166},
  year={1999}
}

@article{xi2025rise,
  title={The rise and potential of large language model based agents: A survey},
  author={Xi, Zhiheng and Chen, Wenxiang and Guo, Xin and He, Wei and Ding, Yiwen and Hong, Boyang and Zhang, Ming and Wang, Junzhe and Jin, Senjie and Zhou, Enyu and others},
  journal={Science China Information Sciences},
  volume={68},
  number={2},
  pages={121101},
  year={2025},
  publisher={Springer}
}

@article{driess2023palm,
  title={Palm-e: An embodied multimodal language model},
  author={Driess, Danny and Xia, Fei and Sajjadi, Mehdi SM and Lynch, Corey and Chowdhery, Aakanksha and Wahid, Ayzaan and Tompson, Jonathan and Vuong, Quan and Yu, Tianhe and Huang, Wenlong and others},
  year={2023}
}

@article{lei2021detecting,
  title={Detecting moments and highlights in videos via natural language queries},
  author={Lei, Jie and Berg, Tamara L and Bansal, Mohit},
  journal={Advances in Neural Information Processing Systems},
  volume={34},
  pages={11846--11858},
  year={2021}
}

\appendix

\clearpage
\section{Appendix}

\subsection{Related Works}
To address the fragmented landscape of streaming multimodal models, we unify representative methods in a comparative analysis along two axes: supported input modalities and interaction capabilities. We observe that many works described as ``streaming'' in fact adopt a question-injection protocol, where a query is issued at a predetermined timestamp and the model answers using only the preceding context. As a result, they primarily study long-horizon processing via KV-cache compression and external memory, rather than continuous online interaction with response-timing decisions. In contrast, the few systems that support online streaming interaction typically span all three interaction types. LiveCC is a notable exception: it focuses on fixed-rate real-time narration and therefore does not require deciding when to respond. Moreover, LION-FS, VideoLLM-MoD, and LiveStar mainly introduce efficiency improvements on top of the VideoLLM-online pipeline. Accordingly, we use VideoLLM-online as the representative baseline. Overall, Table~\ref{tab:related_work} shows that our method is the first open-source model to enable full omni-modal streaming while natively supporting proactive response, real-time narration, and reactive QA within a unified framework.

We also summarize commonly used benchmarks for streaming evaluation in Table~\ref{tab:related_bench}. Although these benchmarks are often described as ``streaming'', they target different capabilities, and their coverage is uneven, which motivates us to consolidate them into a unified evaluation protocol.

\subsection{Sensitivity Analysis}
\label{app:sen}
\begin{figure}[htbp]
  \centering
  \includegraphics[width=\columnwidth]{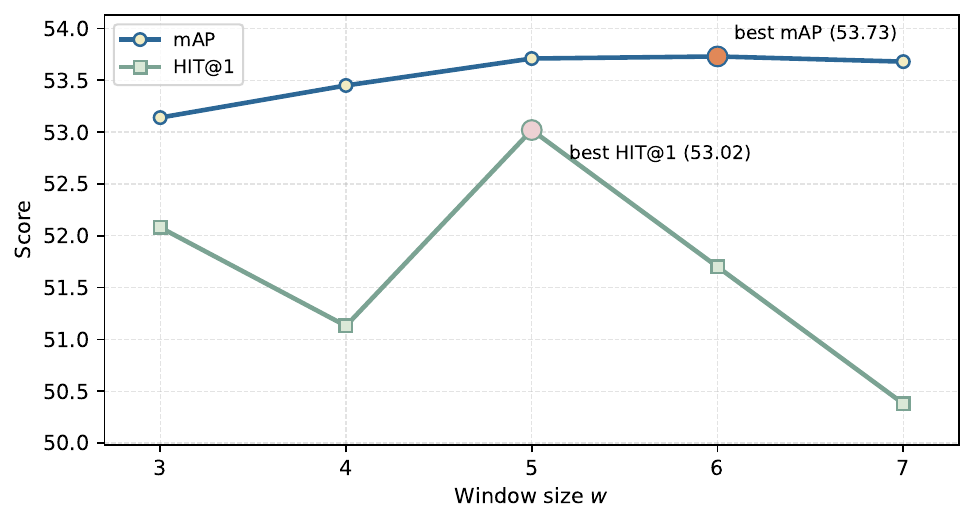}
  \caption{Sensitivity analysis on window size on QVHighlight.}
  \label{fig:qvh}
\end{figure}

Sensitivity analysis confirms robust performance (Figure~\ref{fig:qvh}). In static settings, mAP remains stable while HIT@1 shows only slight sensitivity to variations in the window size. Dynamic tasks exhibit a broad operating regime with smooth degradation, indicating no brittle reliance on specific parameters. Narration is likewise insensitive to speak head probability thresholds, justifying a fixed default without additional tuning (Table~\ref{tab:sen_narration}).

\begin{figure}[htbp]
  \centering
  \includegraphics[width=\columnwidth]{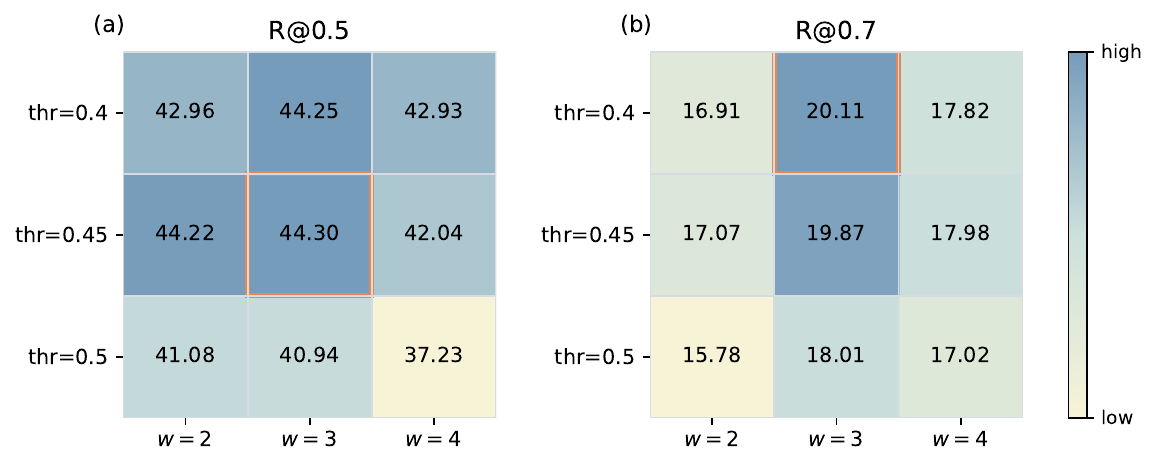}
  \caption{Sensitivity analysis on window size and threshold on Charades-STA.}
  \label{fig:cha}
\end{figure}

\begin{table*}[h]
\centering
\begin{tblr}{
  width = \linewidth,
  colspec = {X[l,1] X[c,0.7] X[c,0.7] X[c,2] X[c,0.7] X[c,0.7] X[c,2]},
  colsep = 1.5pt,
  rowsep = 0.8pt,
  stretch = 0.9,
  rows = {font=\normalsize},
  row{1} = {font=\bfseries\normalsize},
  row{2} = {font=\small},
  cell{1}{1} = {r=2}{c, valign=m},
  cell{1}{2,5} = {c=3}{c, font=\bfseries},
  cells = {valign=m},
  vline{5} = {0.6pt, fg=black!35},
  hline{1,3} = {1pt, solid},
  hline{2} = {2-7}{0.5pt, solid},
  hline{Z} = {1pt, solid},
}
  {Probability Threshold} & YouCook2 & & & OVO-Bench (SSR) & & \\
   & F1 & BERTScore & GPT-Eval & F1 & BERTScore & GPT-Eval \\

  0.965 & 35.36 & 0.82 & 0.52 / 0.28 / 0.31 & \textbf{15.53} & 0.82 & 0.59 / 0.28 / 0.29 \\
  0.970 & 35.05 & 0.82 & 0.50 / 0.29 / 0.33 & 14.54 & 0.83 & \textbf{0.59 / 0.33 / 0.34} \\
  \textbf{0.975} & \textbf{35.21} & \textbf{0.83} & \textbf{0.53 / 0.29 / 0.36} & 14.58 & 0.83 & 0.62 / 0.32 / 0.33 \\
  0.980 & 34.90 & 0.83 & 0.55 / 0.28 / 0.36 & 15.15 & \textbf{0.84} & 0.59 / 0.33 / 0.36 \\
  0.985 & 34.07 & 0.83 & 0.52 / 0.29 / 0.31 & 14.73 & 0.84 & 0.62 / 0.29 / 0.35 \\
\end{tblr}
\caption{Sensitivity analysis of the probability threshold on real-time narration, presenting performance metrics (F1, BERTScore, GPT-Eval) across different triggering thresholds on YouCook2 and OVO-Bench (SSR).}
\label{tab:sen_narration}
\end{table*}

\begin{table*}[h]
\centering
\begin{tblr}{
  width = \linewidth,
  colspec = {X[l,2] X[c,0.7] X[c,0.7] X[c,2] X[c,0.7] X[c,0.7] X[c,2]}, 
  colsep = 1pt,            
  rowsep = 0.8pt,            
  stretch = 0.9,           
  rows = {font=\normalsize},    
  row{1} = {font=\bfseries\normalsize},
  row{2} = {font=\small}, 
  cell{1}{1} = {r=2}{c, valign=m},           
  cell{1}{2,5} = {c=3}{c, font=\bfseries},   
  cells = {valign=m},
  vline{5} = {0.6pt, fg=black!35}, 
  hline{1,3} = {1pt, solid},
  hline{2} = {2-7}{0.5pt, solid},
  hline{Z} = {1pt, solid},
  hline{8} = {0.5pt, fg=tblRuleLight}, 
  row{7} = {bg=tblOurs, font=\normalsize}, 
  cell{8,12}{1} = {c=4}{l, font=\itshape\small, bg=tblSection}, 
  cell{8,12}{5} = {c=3}{bg=tblSection}, 
}
  Method & YouCook2 & & &SSR  & & \\
  & F1 & BERTScore & GPT-Eval & F1 & BERTScore & GPT-Eval \\
  
  TimeChat        & 21.70 & --   & --             & --    & --   & -- \\
  VTG-LLM         & 17.50 & --   & --             & --    & --   & -- \\
  VideoLLM-online & 18.82 & 0.82 & 0.33 / 0.05 / 0.12 & 10.24 & \textbf{0.84} & 0.39 / 0.02 / 0.14 \\
  MMDuet          & 17.81 & 0.83 & 0.31 / 0.26 / 0.12 & 9.02  & 0.79 & 0.42 / 0.29 / 0.21 \\
  
  \textbf{Ours}            & \textbf{35.21} & \textbf{0.83} & \textbf{0.53 / 0.29 / 0.36} & \textbf{14.54} & 0.83 & \textbf{0.59 / 0.33 / 0.34} \\
  - Ablation Study & & & & & & \\
  Mixed Training  & 31.42 & 0.81 & 0.47 / 0.30 / 0.24 & 8.88  & 0.80 & 0.52 / 0.34 / 0.13 \\
  w/o Speak Head   & 9.25    & 0.79   & 0.32 / 0.32 / 0.09            & 3.39    & 0.77   & 0.41 / 0.30 / 0.08 \\
  $K=1$           & 34.43    & 0.82   & 0.51 / 0.30 / 0.31             & 9.64    & 0.78   & 0.49 / 0.24 / 0.22 \\
  
  - Sensitivity Analysis & & & & & & \\
  $w_{pos}=2$     & 27.82 & 0.83 & 0.62 / 0.27 / 0.46 & 10.38 & 0.57 & 0.63 / 0.21 / 0.29 \\
  $w_{pos}=4$     & 35.55 & 0.81 & 0.64 / 0.27 / 0.49 & 13.48 & 0.75 & 0.54 / 0.20 / 0.28 \\
\end{tblr}
\caption{Streaming narration results on YouCook2 and OVO-Bench (SSR). We report F1 for temporal window alignment, and use BERTScore and GPT-4o scores to assess narration quality.}
\label{tab:narration_full}
\end{table*}
\subsection{Evaluation Details}
\label{app:eval_detail}
We specify evaluation protocols for our streaming interaction tasks. For PO, we preprocess each sample to replicate the original benchmark by cropping the video to the annotated ask time and injecting the question at that timestamp, ensuring strictly causal temporal ordering. While Streaming VLM baselines take text prompts, our model takes a speech rendering of the same text to benchmark native multimodal processing. For streaming baselines (e.g., VideoLLM-Online and MMDuet), we record the first-response timestamp and report accuracy as the fraction of samples whose first-response time is within ±2 seconds of the annotated ground-truth time. For REC, a model gets one point if its chosen response time falls within the annotated event interval. Each segment is evaluated once, and we report the micro success rate. For CRR, we award one point if the first response after the ask time occurs after the annotated clue time, validating that the model waits for necessary visual evidence before answering. Hyperparameters were determined via validation sets as follows: QVHighlight window size = 5; Charades-STA window size = 3 with threshold = 0.45; PA window size = 5 with threshold = 0.5; PO window size = 4 with threshold = 0.2; REC window size = 2 with threshold = 0.7; CRR window size = 2 with threshold = 0.7; YouCook2 threshold = 0.975; SSR threshold = 0.97.

Due to space constraints, in the main table we report only the average score on the narration task, computed as the mean of the three GPT-4o–based evaluation dimensions. We present the full breakdown in Table~\ref{tab:narration_full}.

\begin{table*}[htbp]
\centering
\footnotesize
\setlength{\tabcolsep}{3pt}
\renewcommand{\arraystretch}{1.26}

\begin{threeparttable}
\renewcommand{\tabularxcolumn}[1]{m{#1}}
\begin{tabularx}{\textwidth}{L{2.7cm} C{1.1cm} C{1.3cm} C{1.3cm} C{1.35cm} Y}
\toprule
\rowcolor{softblue}
\textbf{Method} & \textbf{Inputs} & \textbf{Alert} & \textbf{Narration} & \textbf{Reactive QA} & \textbf{Description} \\
\midrule

\rowcolor{grpA}
Moshi \cite{defossez2024moshi} & A & \no & \no & \yes & A full-duplex speech--text model enabling low-latency, real-time voice dialogue without vision. \\
\rowcolor{grpA}
Voila \cite{shi2025voila} & A & \no & \no & \yes & Real-time voice-language model with expressive role-play and vocal styles. \\

\rowcolor{grpB}
VideoStreaming \cite{qian2024streaming} & T+V & \no & \no & \yes & Streaming framework with hierarchical memory for coherent long-video understanding. \\
\rowcolor{grpB}
Flash-VStream \cite{zhang2024flash} & T+V & \no & \no & \yes & Real-time system with lightweight memory for low-latency long video processing. \\
\rowcolor{grpB}
VideoLLaMB \cite{wang2024videollamb} & T+V & \no & \no & \yes & Long-context model with recurrent memory bridges to propagate information over time. \\
\rowcolor{grpB}
InternLM-XComposer2.5 \cite{zhang2024internlm} & T+V & \no & \no & \yes & Multimodal streaming system with layered memory for long-term video--audio interaction. \\
\rowcolor{grpB}
StreamChat \cite{xiong2025streaming} & T+V & \no & \no & \yes & Streaming agent using hierarchical memory to sustain long-context, multi-round dialogue. \\
\rowcolor{grpB}
StreamBridge \cite{wang2025streambridge} & T+V & \no & \no & \yes & Plug-and-play buffer turning offline Video-LLMs into proactive streaming assistants. \\
\rowcolor{grpB}
CogStream \cite{zhao2025cogstream} & T+V & \no & \no & \yes & QA framework keeping only memory-critical context for efficient reasoning. \\

\rowcolor{grpC}
ReKV \cite{di2025streaming} & T+V & \no & \no & \yes & Retrieves historical KV-cache as in-context memory instead of re-encoding past frames. \\
\rowcolor{grpC}
LiveVLM \cite{ning2025livevlm}& T+V & \no & \no & \yes & Optimizes streaming KV-cache update/retrieval for efficient long-horizon processing. \\
\rowcolor{grpC}
StreamAgent \cite{yang2025streamagent}& T+V & \no & \no & \yes & Uses KV-based temporal memory to anticipate events and respond proactively. \\
\rowcolor{grpC}
StreamingVLM \cite{xu2025streamingvlm}& T+V & \no & \no & \yes & Manages rolling KV-cache to support infinite video streams under bounded computation. \\
\rowcolor{grpC}
StreamKV \cite{chen2025streamkv}& T+V & \no & \no & \yes & Segments and compresses KV-cache, keeping only salient past segments within budget. \\

\rowcolor{grpD}
LiveCC \cite{chen2025livecc}& T+V & \no & \yes & \no & Trained with continuous speech transcription for real-time narration of long videos. \\

\rowcolor{grpD}
VideoLLM-online \cite{chen2024videollm}& T+V & \yes & \yes & \yes & Online model for temporally aligned dialogue via a LIVE training framework. \\
\rowcolor{grpD}
AssistPDA \cite{yang2025assistpda}& T+V & \yes & \yes & \yes & Surveillance assistant unifying anomaly prediction and interactive analysis. \\
\rowcolor{grpD}
LION-FS \cite{li2025lion}& T+V & \yes & \yes & \yes & Uses fast--slow thinking with selective tokenization for efficient streaming. \\
\rowcolor{grpD}
VideoLLM-MoD~\cite{wu2024videollm}& T+V & \yes & \yes & \yes & Enables efficient streaming by letting each layer skip a subset of tokens directly. \\
\rowcolor{grpD}
LiveStar \cite{yang2025livestar}& T+V & \yes & \yes & \yes & Uses perplexity-based timing and streaming-aware attention for proactive understanding. \\
\rowcolor{grpD}
MMDuet \cite{wang2024videollm}& T+V & \yes & \yes & \yes & Adopts a video--text duet format to insert replies during continuous playback. \\
\rowcolor{grpD}
Dispider \cite{qian2025dispider}& T+V & \yes & \yes & \yes & Disentangles perception, decision, and reaction for asynchronous responses. \\

\rowcolor{grpE}
MiniCPM-o 2.6 \cite{yao2024minicpm}& T+V+A & \no & \no & \yes & Omni model supporting low-latency streaming speech over text, audio, and video. \\
\rowcolor{grpE}
Qwen2.5-Omni \cite{xu2025qwen2}& T+V+A & \no & \no & \yes & Dense Thinker--Talker model with streaming encoders for real-time perception. \\
\rowcolor{grpE}
Qwen3-Omni \cite{Qwen3-Omni}& T+V+A & \no & \no & \yes & Natively omni-modal MoE architecture for high-concurrency streaming generation. \\
\rowcolor{grpE}
Stream-Omni \cite{zhang2025stream}& T+V+A & \no & \no & \yes & Efficient modality alignment enabling speech interaction grounded on visual inputs. \\
\rowcolor{grpE}
ViSpeak \cite{fu2025vispeak}& T+V+A & \yes & \no & \yes & Vision-centric framework producing instruction-like feedback from evolving streams. \\

\rowcolor{grpOurs}
\textbf{ROMA (Ours)} & T+V+A & \yes & \yes & \yes & \textbf{End-to-end streaming OLLM processing time-aligned chunks for proactive interaction.} \\
\bottomrule
\end{tabularx}

\caption{Comparison of different streaming multimodal methods. \textbf{Note:} T=Text, V=Visual, A=Audio. \yes~supports the ability, \no~does not.}
\label{tab:related_work}
\end{threeparttable}
\end{table*}

\subsection{Data Construction Details}
\label{app:data}
\paragraph{Proactive Data Processing} Since the original temporal annotations in DiDeMo and Charades-STA are often coarse, simply using them for streaming supervision introduces noise. We therefore re-annotate event windows using Doubao-Seed-1.6-thinking~\cite{seed2025seed1} to obtain precise start and end timestamps. For timing supervision, we label every second within the refined ground-truth event window as a positive trigger, ensuring the model learns robust event sensitivity.

\paragraph{Narration Data Processing} 
Raw videos often contain unlabeled gaps that induce hallucination during training. We mitigate this by excising these intervals and concatenating annotated segments into continuous, semantically dense streams with recalibrated timestamps. For timing supervision, we discard the broad-window labeling of prior work (e.g., MMDuetIT) in favor of strict transition-based triggering. This yields precise supervision for incremental narration via multi-turn SFT.

\paragraph{Audio Query Synthesis} To simulate real-world interaction, we synthesize text queries via TTS and overlay them onto original audio tracks. We strictly align these spoken queries with streaming units to enforce audio-driven instruction following. 
\subsection{Implementation Details}
\label{app:impl}
\paragraph{Training Configuration.} We sample videos at 2 FPS and resize frames to a maximum of 65,536 pixels. The model is trained using LLaMA-Factory~\cite{zheng2024llamafactory} with a sequence length of 32K on 32 H20 GPUs (using a global batch size of 512). Proactive samples are specifically formatted as multi-turn dialogues to handle multiple triggers within a single stream.

\paragraph{Streaming Decoding Logic.} Following the pipelined setting, if a response exceeds the 25-token budget, we append an \tok{endoftext} (<eot>) token to signal an unfinished utterance. Decoding resumes in the subsequent segment and terminates only when \tok{im\_{end}} is generated.

\subsection{Case Study}
\begin{figure*}[h]
  \centering
  \includegraphics[width=\textwidth]{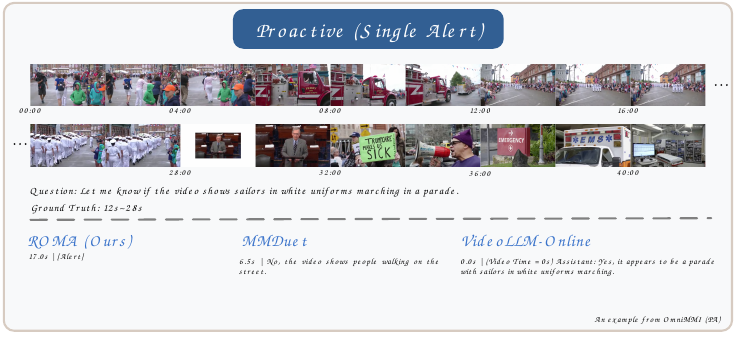}
  \caption{Qualitative comparison on the single-alert proactive task. While MMDuet and VideoLLM-Online exhibit premature triggering and hallucination before the target event appears, \ours accurately accumulates visual evidence to release a precise alert at 17.0s, aligning with the ground truth interval (12s--28s).}
  \label{fig:case_pro_one}
\end{figure*}

\begin{figure*}[h]
  \centering
  \includegraphics[width=\textwidth]{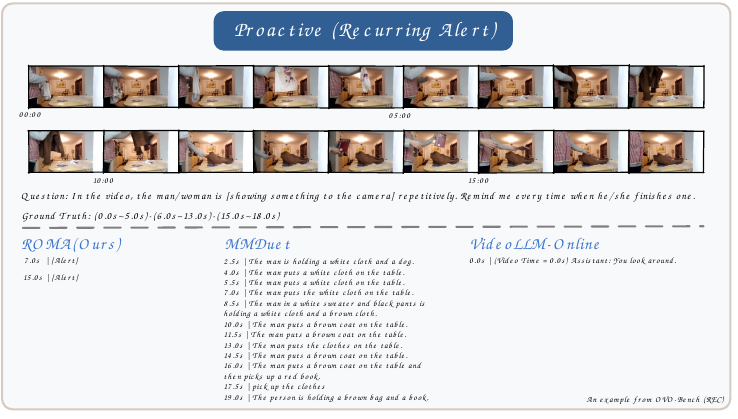}
  \caption{Qualitative comparison on the recurring-alert task. While MMDuet suffers from continuous over-generation without distinguishing event boundaries, \ours effectively tracks the repetitive action, releasing distinct alerts at 7.0s and 15.0s to capture the recurring instances.}
  \label{fig:case_pro_multi}
\end{figure*}

\paragraph{Event-Triggered Alert}
We present two event-triggered alert cases: one where the target event occurs only once (Figure~\ref{fig:case_pro_one}), and another where it recurs multiple times (Figure~\ref{fig:case_pro_multi}). Compared with several representative VideoLLMs, our model triggers at more accurate times.

\begin{figure*}[h]
  \centering
  \includegraphics[width=\textwidth]{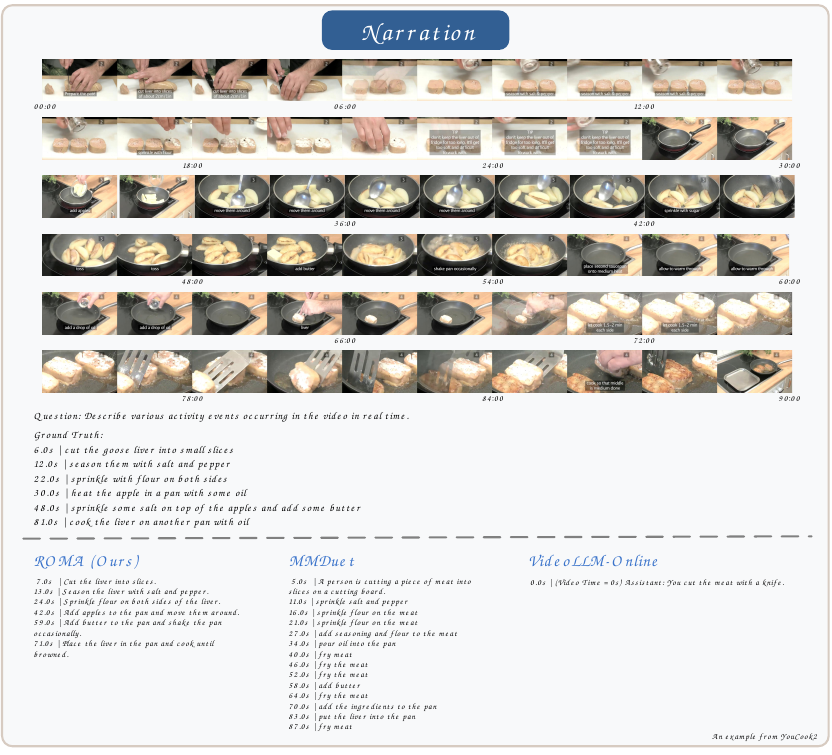}
  \caption{Qualitative comparison on the real-time narration task. While MMDuet suffers from severe repetition and redundant over-generation, \ours effectively tracks the procedural evolution, generating concise, time-aligned descriptions that correspond strictly to the distinct ground truth events.}
  \label{fig:case_yc2}
\end{figure*}

\paragraph{Narration}
In the narration task, the model must choose when to speak during a long streaming video and provide concise summaries of events observed so far without access to future content. As shown in Figure~\ref{fig:case_yc2}, compared with VideoLLMs, our outputs are more succinct and our response timings align more closely with key event boundaries, leading to more accurate online narration.

\begin{figure*}[h]
  \centering
  \includegraphics[width=\textwidth]{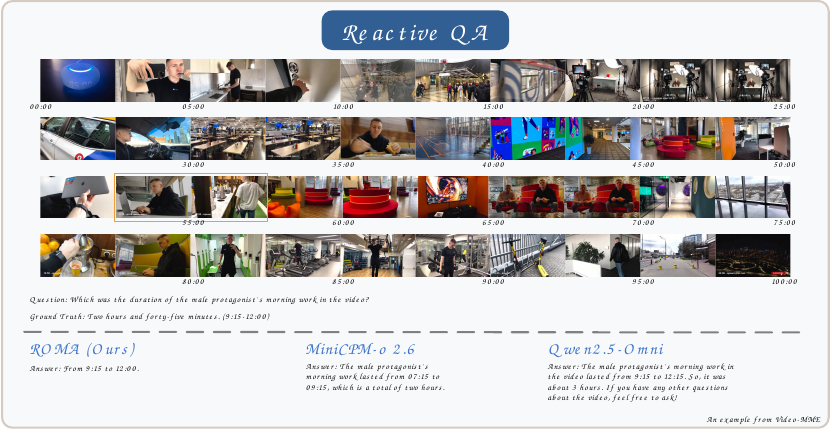}
  \caption{Qualitative comparison on the reactive QA task. While baseline models suffer from temporal misalignment or hallucinated intervals when querying specific activity durations, \ours accurately retrieves the exact start and end timestamps (9:15--12:00) to derive the correct answer.}
  \label{fig:case_turn}
\end{figure*}

\paragraph{Reactive QA}
With audio queries in the reactive QA setting (Figure~\ref{fig:case_turn}), \ours correctly localizes the relevant segment in the long video and extracts the key visual evidence. In contrast, MiniCPM-o misidentifies the segment, while Qwen2.5-Omni often responds with unnecessary follow-up questions.

\subsection{Evaluation Prompt}
LLM-as-a-judge is a widely adopted paradigm for scalable evaluation, given its strong alignment with human preferences~\cite{zheng2023judging}. Accordingly, we employ GPT-4o as a reliable scorer for our open-ended tasks. Detailed prompts are provided in Figures~\ref{fig:open-eval-prompt} and \ref{fig:prompt-narration}.

\begin{figure*}[t]
\centering
\begin{promptbox}{Evaluation Prompt for Open-Ended Audio-Query Assessment}
\begin{minipage}{\tcbtextwidth}
\begin{lstlisting}[style=PromptStyle]
"""
You are an AI assistant tasked with evaluating whether a response matches the correct answer to a given question.

## Evaluation Rules
(1) Output 1 if the response matches the answer exactly or with synonymous/ equivalent wording.
- Synonyms, paraphrases, or different surface forms of the same meaning count as matches.
- Responses containing the correct answer with irrelevant details count as matches.
- Responses providing sufficient information to infer the correct answer count as matches.

(2) Output 0 if the response is incorrect, contradictory, or completely irrelevant to the question.
- If the answer and the response address different topics, or if the response does not answer the question.
- If the response introduces additional details that change the meaning of the answer, mark as 0.

### Examples 
Example 1: 
Question: What is the genre of this video? 
Answer: It is a news report that introduces the history behind Christmas decorations. 
Response: It's a Christmas-themed video, filled with festive decorations and a warm, cozy atmosphere. It really captures that classic holiday spirit. What do you think? 
Your output: 0 

Example 2: 
Question: How many birds are above the fireplace? 
Answer: 2. 
Response: One bird is above the fireplace, and another is below. 
Your output: 1

Your Turn:
Question: {question}
Answer: {ground_truth}
Response: {prediction}
Your output:
"""
\end{lstlisting}
\end{minipage}
\end{promptbox}
\caption{Full prompt provided to GPT-4o for open-ended evaluation with audio queries on Video-MME and EgoSchema.}

\label{fig:open-eval-prompt}
\end{figure*}

\begin{figure*}[t]
\centering
\begin{promptbox}{Evaluation Prompt for Narration Task Evaluation}
\begin{minipage}{\tcbtextwidth}
\begin{lstlisting}[style=PromptStyle]
"""
You are an expert evaluator for video narration quality. Your task is to compare
a reference description of a video (ground truth) with a model-generated description
for the same video, and output THREE scores between 0 and 1.

You must consider the model response as a SINGLE long story (it may contain multiple
sentences describing different moments in the video).

IMPORTANT: Higher scores are always better.

Definitions:

1. coherence (story coherence):
   - How internally coherent and well-structured is the model-generated story by itself?
   - Does it read like a reasonable, temporally plausible sequence of actions and states?
   - Penalize contradictions, abrupt jumps, and incoherent, rambling structure.
   - 1.0 = very coherent and well-structured; 0.0 = completely incoherent.

2. alignment (semantic alignment with ground truth):
   - How well does the model-generated story capture the key actions and steps in the ground truth?
   - Consider whether important actions/events are present, correctly described, and roughly in a reasonable order.
   - Hallucinated major steps that clearly do not appear in the ground truth should reduce this score.
   - 1.0 = almost all key content in GT is covered with correct semantics; 0.0 = almost completely unrelated.

3. conciseness (relevant non-redundancy / brevity):
   - This score measures whether the model response is concise GIVEN IT IS RELEVANT to the ground truth.
   - If the model response is largely unrelated to the ground truth (low semantic overlap, wrong topic, ignores the video), conciseness MUST be near 0, even if the response is short.
   - Penalize heavy repetition of similar sentences, long irrelevant digressions, and obvious padding.
   - However, do NOT penalize necessary detail that genuinely helps describe the steps.
   - 1.0 = succinct, minimal redundancy while preserving essential details;
     0.0 = extremely repetitive / rambling / full of irrelevant filler / irrelevant with the groundtruth.
     

Empty or meaningless model responses (or responses that ignore the task) should receive low scores,
typically near 0 for all dimensions.

Output format (VERY IMPORTANT):
- You MUST output valid JSON with exactly the following keys:
  {"coherence": <float>, "alignment": <float>, "conciseness": <float>}
- Each value must be a number between 0 and 1 (inclusive).
- Do NOT output any extra text or explanation.
"""
\end{lstlisting}
\end{minipage}
\end{promptbox}
\caption{Prompt used to instruct GPT-4o to evaluate narration quality along three criteria: story coherence, alignment with ground truth, and conciseness.}

\label{fig:prompt-narration}
\end{figure*}

\end{document}